\documentclass{article} % For LaTeX2e
\usepackage{iclr2027_conference,times}
\ifdefined\XeTeXversion
  \usepackage[no-math]{fontspec}
\fi
\usepackage{amsmath,amsfonts,bm}

\def\eqref#1{equation~\ref{#1}}
\def\1{\bm{1}}

\DeclareMathAlphabet{\mathsfit}{\encodingdefault}{\sfdefault}{m}{sl}
\SetMathAlphabet{\mathsfit}{bold}{\encodingdefault}{\sfdefault}{bx}{n}

\usepackage{hyperref}
\usepackage{url}
\usepackage{graphicx}
\usepackage{booktabs}
\usepackage{multirow}
\usepackage{amsmath,amssymb,amsthm}
\usepackage{xcolor}
\usepackage{subcaption}
\usepackage{wrapfig}
\usepackage{enumitem}
\usepackage{tikz}
\usepackage{algorithm}
\usepackage{algpseudocode}
\algrenewcommand\algorithmicrequire{\textbf{Input:}}
\algrenewcommand\algorithmicensure{\textbf{Output:}}

\newcommand{\ours}{OpenAttention}
\newcommand{\qgsfull}{Query Geometry Selector}
\newcommand{\qgcfull}{Query Geometry Clustering}
\newcommand{\qgc}{QGC}
\newcommand{\qgs}{QGS}
\newcommand{\mgate}{ErrorGate}
\newcommand{\oursflash}{\ours{}.flash}

\newcommand{\vatt}{vAttention}
\newcommand{\mpig}{MagicPIG}

\newtheorem{proposition}{Proposition}
\newtheorem{lemma}{Lemma}
\newtheorem{corollary}{Corollary}

\newcommand{\tablongbenchrows}{%
dense & 36.08 & 26.61 \\
HashAttention & 35.34 & 22.09 \\
\vatt{} & 31.05 & 16.33 \\
\ours{} (HashAttention selector, static-metric partition) & 35.95 & 26.99 \\
\ours{} (static metric only: selector and partition) & 35.53 & 26.80 \\
\ours{} (static $+$ instruction tokens as question) & 34.68 & 26.93 \\
}
\newcommand{\tablbvtwo}{%
dense & 30.00 \\
HashAttention & 23.30 \\
\ours{} (\qgs{}) & 33.30 \\
}

\newcommand{\tabvattprobe}{%
\multicolumn{3}{l}{\emph{target read 1\%}: 14 of 40 configurations within $1.05\times$ the target} \\
$\epsilon{=}\delta{=}0.3$, nominal 1.0\%, sampling share 0.50 \textbf{(chosen)} & 1.00\% & 0.470 \\
$\epsilon{=}\delta{=}0.2$, nominal 0.8\%, sampling share 0.50 & 0.90\% & 0.579 \\
$\epsilon{=}\delta{=}0.1$, nominal 0.4\%, sampling share 0.25 & 0.83\% & 0.628 \\
$\epsilon{=}\delta{=}0.05$, nominal 0.4\%, sampling share 0.25 (none fits: smallest read) & 3.16\% & 0.489 \\
$\epsilon{=}\delta{=}0.025$, nominal 0.4\%, sampling share 0.25 (none fits: smallest read) & 9.32\% & 0.383 \\
\midrule
\multicolumn{3}{l}{\emph{target read 3\%}: 25 of 60 configurations within $1.05\times$ the target} \\
$\epsilon{=}\delta{=}0.3$, nominal 3.0\%, sampling share 0.50 \textbf{(chosen)} & 3.00\% & 0.339 \\
$\epsilon{=}\delta{=}0.2$, nominal 2.4\%, sampling share 0.50 & 2.67\% & 0.360 \\
$\epsilon{=}\delta{=}0.1$, nominal 1.8\%, sampling share 0.25 & 3.00\% & 0.369 \\
$\epsilon{=}\delta{=}0.05$, nominal 1.2\%, sampling share 0.25 (none fits: smallest read) & 5.74\% & 0.329 \\
$\epsilon{=}\delta{=}0.025$, nominal 1.2\%, sampling share 0.25 (none fits: smallest read) & 13.84\% & 0.246 \\
\midrule
\multicolumn{3}{l}{\emph{target read 5\%}: 26 of 60 configurations within $1.05\times$ the target} \\
$\epsilon{=}\delta{=}0.3$, nominal 5.0\%, sampling share 0.50 \textbf{(chosen)} & 4.97\% & 0.275 \\
$\epsilon{=}\delta{=}0.2$, nominal 5.0\%, sampling share 0.25 & 5.23\% & 0.278 \\
$\epsilon{=}\delta{=}0.1$, nominal 3.0\%, sampling share 0.25 & 4.75\% & 0.299 \\
$\epsilon{=}\delta{=}0.05$, nominal 2.0\%, sampling share 0.25 (none fits: smallest read) & 7.32\% & 0.270 \\
$\epsilon{=}\delta{=}0.025$, nominal 2.0\%, sampling share 0.25 (none fits: smallest read) & 16.33\% & 0.208 \\
\midrule
\multicolumn{3}{l}{\emph{target read 10\%}: 32 of 75 configurations within $1.05\times$ the target} \\
$\epsilon{=}\delta{=}0.3$, nominal 10.2\%, sampling share 0.25 \textbf{(chosen)} & 10.03\% & 0.204 \\
$\epsilon{=}\delta{=}0.2$, nominal 10.0\%, sampling share 0.25 & 10.44\% & 0.205 \\
$\epsilon{=}\delta{=}0.1$, nominal 6.0\%, sampling share 0.50 & 9.71\% & 0.220 \\
$\epsilon{=}\delta{=}0.05$, nominal 4.0\%, sampling share 0.25 & 10.45\% & 0.208 \\
$\epsilon{=}\delta{=}0.025$, nominal 4.0\%, sampling share 0.25 (none fits: smallest read) & 20.12\% & 0.170 \\
}
\newcommand{\tabvattfixed}{%
1\% & 37.36 & 36.29 & 1.0\% \\
3\% & 57.15 & 55.74 & 3.0\% \\
5\% & 68.70 & 68.86 & 5.0\% \\
10\% & 82.26 & 83.08 & 10.0\% \\
}
\newcommand{\tablenrows}{%
\multirow{5}{*}{16K}
& dense & 100\% & 87.14 & 86.0 & 51.0 & 93.2 & 91.0 & 95.0 & 98.0 & 95.8 \\
& \mpig{} & 3.0\% & 46.91 & 49.0 & 32.0 & 83.2 & 79.7 & 14.0 & 0.0 & 70.5 \\
& HashAttention & 3.0\% & 67.41 & 84.0 & 53.0 & 78.4 & 83.0 & 62.0 & 20.0 & 91.5 \\
& \vatt{} & 3.0\% & 52.69 & 60.0 & 41.0 & 73.4 & 83.7 & 21.0 & 1.0 & 88.8 \\
& \textbf{\ours{} (ours)} & 3.0\% & 87.49 & 86.0 & 54.0 & 94.0 & 89.7 & 95.0 & 98.0 & 95.8 \\
\midrule
\multirow{5}{*}{32K}
& dense & 100\% & 87.75 & 83.0 & 50.0 & 95.2 & 93.3 & 96.0 & 100.0 & 96.8 \\
& \mpig{} & 3.2\% & 51.38 & 46.0 & 44.0 & 83.0 & 81.7 & 16.0 & 1.0 & 88.0 \\
& HashAttention & 3.0\% & 63.87 & 67.0 & 48.0 & 80.4 & 87.7 & 54.0 & 16.0 & 94.0 \\
& \vatt{} & 3.0\% & 55.63 & 58.0 & 38.0 & 77.6 & 89.3 & 34.0 & 3.0 & 89.5 \\
& \textbf{\ours{} (ours)} & 3.0\% & 87.38 & 83.0 & 51.0 & 96.0 & 91.7 & 95.0 & 99.0 & 96.0 \\
\midrule
\multirow{5}{*}{64K}
& dense & 100\% & 85.49 & 81.0 & 49.0 & 90.6 & 88.3 & 97.0 & 98.0 & 94.5 \\
& \mpig{} & 3.2\% & 45.73 & 41.0 & 34.0 & 73.6 & 77.0 & 12.0 & 0.0 & 82.5 \\
& HashAttention & 3.0\% & 49.65 & 64.0 & 36.0 & 60.6 & 86.7 & 15.0 & 0.0 & 85.2 \\
& \vatt{} & 3.0\% & 54.79 & 67.0 & 37.0 & 83.0 & 85.0 & 19.0 & 3.0 & 89.5 \\
& \textbf{\ours{} (ours)} & 3.0\% & 84.82 & 81.0 & 51.0 & 90.8 & 81.7 & 97.0 & 98.0 & 94.2 \\
\midrule
\multirow{5}{*}{96K}
& dense & 100\% & 78.01 & 78.0 & 47.0 & 61.0 & 90.3 & 84.0 & 89.0 & 96.8 \\
& \mpig{} & 3.0\% & 33.19 & 31.0 & 23.0 & 57.6 & 57.0 & 3.0 & 0.0 & 60.8 \\
& HashAttention & 3.0\% & 27.99 & 14.0 & 19.0 & 23.6 & 74.3 & 0.0 & 0.0 & 65.0 \\
& \vatt{} & 3.0\% & 40.20 & 34.0 & 26.0 & 50.2 & 85.7 & 6.0 & 0.0 & 79.5 \\
& \textbf{\ours{} (ours)} & 3.0\% & 78.10 & 79.0 & 47.0 & 62.8 & 87.7 & 84.0 & 89.0 & 97.2 \\
}
\newcommand{\tablenavg}{%
dense & 87.14 & 87.75 & 85.49 & 78.01 \\
\mpig{} & 46.91 & 51.38 & 45.73 & 33.19 \\
HashAttention & 67.41 & 63.87 & 49.65 & 27.99 \\
\vatt{} & 52.69 & 55.63 & 54.79 & 40.20 \\
\textbf{\ours{} (ours)} & 87.49 & 87.38 & 84.82 & 78.10 \\
}

\newcommand{\tablatencyrows}{%
dense attention (fused SDPA) & -- & 0.06 & 1.20 & 0.06 & 1.22 \\
HashAttention & \xmark & 0.12 & 0.54 & 0.11 & 1.65 \\
\vatt{} (lower bound) & \xmark & 0.37 & 0.67 & 0.35 & 1.75 \\
\mpig{} (lower bound) & \xmark & 0.13 & 0.55 & 0.11 & 1.65 \\
\ours{}, unfused gather & \xmark & 0.29 & 0.46 & 0.27 & 1.26 \\
\textbf{\oursflash{}} & \cmark & \textbf{0.05} & \textbf{0.12} & \textbf{0.07} & \textbf{0.24} \\
}
\newcommand{\tabdensityours}{%
100\% & dense & 100\% & 87.04 & 79.0 & 50.0 & 95.2 & 93.3 & 95.0 & 100.0 & 96.8 \\
\midrule
\multirow{4}{*}{10\%}
& \mpig{} & 10.0\% & 52.52 & 51.0 & 39.0 & 90.2 & 88.8 & 30.0 & 3.5 & 65.1 \\
& HashAttention & 10.0\% & 81.23 & 70.5 & 45.5 & 87.9 & 89.8 & 88.0 & 91.0 & 95.9 \\
& \vatt{} & 10.0\% & 83.08 & 73.0 & 46.0 & 91.3 & 92.2 & 92.5 & 91.0 & 95.6 \\
& \textbf{\ours{} (ours)} & 10.0\% & \textbf{87.08} & 79.0 & 51.0 & 95.3 & 92.7 & 95.0 & 100.0 & 96.6 \\
\midrule
\multirow{4}{*}{5\%}
& \mpig{} & 5.1\% & 34.42 & 36.0 & 33.5 & 71.5 & 88.3 & 2.5 & 0.0 & 9.1 \\
& HashAttention & 5.0\% & 71.55 & 68.0 & 44.5 & 81.9 & 87.3 & 72.5 & 50.5 & 96.1 \\
& \vatt{} & 5.0\% & 68.86 & 67.5 & 45.0 & 85.8 & 90.5 & 67.0 & 33.5 & 92.8 \\
& \textbf{\ours{} (ours)} & 5.0\% & \textbf{86.82} & 78.5 & 50.5 & 95.8 & 92.2 & 94.5 & 99.5 & 96.8 \\
\midrule
\multirow{4}{*}{3\%}
& \mpig{} & 2.9\% & 24.01 & 24.5 & 26.0 & 34.0 & 81.8 & 0.0 & 0.0 & 1.8 \\
& HashAttention & 3.0\% & 63.48 & 64.0 & 43.0 & 80.9 & 86.8 & 58.5 & 16.0 & 95.1 \\
& \vatt{} & 3.0\% & 55.74 & 55.0 & 36.5 & 79.1 & 88.7 & 38.0 & 3.0 & 89.9 \\
& \textbf{\ours{} (ours)} & 3.0\% & \textbf{86.83} & 78.5 & 51.0 & 95.9 & 92.0 & 94.5 & 99.5 & 96.4 \\
\midrule
\multirow{4}{*}{1\%}
& \mpig{} & 1.1\% & 10.53 & 12.0 & 13.5 & 0.7 & 47.5 & 0.0 & 0.0 & 0.0 \\
& HashAttention & 1.0\% & 48.22 & 44.0 & 41.5 & 73.3 & 78.5 & 12.5 & 0.0 & 87.8 \\
& \vatt{} & 1.0\% & 36.29 & 24.0 & 23.0 & 62.6 & 73.7 & 1.5 & 0.0 & 69.2 \\
& \textbf{\ours{} (ours)} & 1.0\% & \textbf{86.30} & 77.5 & 50.0 & 96.0 & 93.5 & 95.0 & 97.0 & 95.1 \\
}

\newcommand{\OURSTEN}{87.08}
\newcommand{\OURSFIVE}{86.82}
\newcommand{\OURSTHREE}{86.83}
\newcommand{\OURSONE}{86.30}

\newcommand{\PIPEHITHATONE}{0.5\%}

\newcommand{\PIPENROWS}{137,216}
\newcommand{\tabdenstune}{%
10\% & 512 & 2493 & 32 & 23\% & 86.59 & 78.0 & 49.5 & 95.2 & 92.7 & 95.0 & 99.5 & 96.2 \\
10\% & 256 & 2877 & 32 & 12\% & 86.85 & 78.0 & 50.5 & 95.0 & 94.2 & 94.5 & 99.5 & 96.2 \\
5\% & 256 & 1238 & 32 & 23\% & 86.51 & 76.5 & 51.0 & 94.9 & 93.2 & 94.5 & 99.5 & 96.0 \\
5\% & 192 & 1335 & 32 & 18\% & 86.98 & 79.0 & 50.5 & 95.1 & 93.5 & 95.0 & 99.5 & 96.2 \\
5\% & 128 & 1431 & 32 & 12\% & 86.83 & 78.5 & 50.0 & 95.2 & 94.3 & 95.0 & 99.5 & 95.2 \\
10\% & 384 & 2685 & 32 & 18\% & 86.60 & 77.5 & 50.0 & 95.1 & 93.0 & 95.0 & 99.5 & 96.1 \\
3\% & 256 & 583 & 32 & 39\% & 86.69 & 78.0 & 50.5 & 95.4 & 92.7 & 95.0 & 99.5 & 95.8 \\
3\% & 192 & 679 & 16 & 29\% & 86.63 & 79.0 & 49.0 & 93.5 & 94.2 & 95.5 & 99.5 & 95.8 \\
3\% & 192 & 679 & 32 & 29\% & 86.80 & 78.5 & 50.5 & 94.8 & 93.3 & 95.0 & 99.5 & 96.0 \\
3\% & 128 & 775 & 32 & 20\% & 86.60 & 77.5 & 49.5 & 95.3 & 93.5 & 95.5 & 99.5 & 95.4 \\
}
\newcommand{\KVCALONEstreamingllm}{28.00}
\newcommand{\KVCPAIRstreamingllm}{28.28}

\newcommand{\KVCALONEsnapkv}{69.98}
\newcommand{\KVCPAIRsnapkv}{69.49}

\newcommand{\KVCALONEpyramidkv}{70.34}
\newcommand{\KVCPAIRpyramidkv}{70.14}
\newcommand{\KVCDELTAAVG}{-0.14}

\newcommand{\KVCBUDGET}{2048}

\newcommand{\tabkvcompress}{%
dense attention & \textbf{87.04} & 79.0 & 50.0 & 95.2 & 93.3 & 95.0 & 100.0 & 96.8 \\
\midrule
StreamingLLM (2048) & \textbf{28.00} & 50.5 & 37.5 & 5.3 & 88.3 & 5.0 & 5.0 & 4.4 \\
\quad + \ours{} at 10\% & \textbf{28.28} & 52.0 & 37.5 & 5.4 & 88.7 & 5.0 & 5.0 & 4.4 \\
\midrule
SnapKV (2048) & \textbf{69.98} & 77.5 & 49.0 & 88.6 & 82.2 & 93.5 & 3.0 & 96.1 \\
\quad + \ours{} at 10\% & \textbf{69.49} & 76.0 & 48.0 & 87.7 & 83.5 & 93.5 & 2.5 & 95.2 \\
\midrule
PyramidKV (2048) & \textbf{70.34} & 78.0 & 49.5 & 90.6 & 82.5 & 93.5 & 2.0 & 96.2 \\
\quad + \ours{} at 10\% & \textbf{70.14} & 76.5 & 49.5 & 89.5 & 83.3 & 94.0 & 2.5 & 95.6 \\
}
\newcommand{\CDNROWS}{137{,}216}

\newcommand{\PARTERRADJ}{0.041}
\newcommand{\PARTDISPADJ}{1.83}
\newcommand{\PARTERRKEY}{0.040}
\newcommand{\PARTDISPKEY}{1.67}
\newcommand{\PARTERROURS}{0.030}
\newcommand{\PARTDISPOURS}{1.05}

\newcommand{\ESTRHODF}{0.92}
\newcommand{\ESTRHOTRUE}{0.92}

\newcommand{\GATEOVDT}{91\%}
\newcommand{\GATEOVFT}{97\%}
\newcommand{\TOPEIGHTERR}{84\%}
\newcommand{\TOPEIGHTTOPK}{95\%}
\newcommand{\NTOPBIGFRAC}{4\%}
\newcommand{\NTOPBIGSHARE}{69\%}
\newcommand{\NTOPZEROFRAC}{84\%}
\newcommand{\NTOPZEROSHARE}{12\%}
\newcommand{\SELNREQ}{21}

\newcommand{\tabablationfiverows}{%
\xmark & \xmark & \xmark & 67.35 & 62.0 & 45.5 & 84.0 & 91.7 & 63.0 & 31.5 & 93.8 \\
\cmark & \xmark & \xmark & 77.91 & 74.5 & 45.5 & 91.8 & 92.8 & 82.5 & 63.0 & 95.2 \\
\cmark & \cmark & \xmark & 85.90 & 78.5 & 49.0 & 95.6 & 92.2 & 94.0 & 96.0 & 96.0 \\
\cmark & \xmark & \cmark & 86.73 & 78.0 & 51.0 & 95.6 & 92.0 & 94.5 & 99.5 & 96.5 \\
\cmark & \cmark & \cmark & \textbf{86.82} & 78.5 & 50.5 & 95.8 & 92.2 & 94.5 & 99.5 & 96.8 \\
}

\newcommand{\tabrankrulerows}{%
mean logit (mass) & 85.99 \\
variance & 84.81 \\
\textbf{estimated error (ours)} & \textbf{86.30} \\
}

\newcommand{\cmark}{\checkmark}
\newcommand{\xmark}{$\times$}
\title{OpenAttention: One Percent Is Enough with Tail Proxies}

\usepackage{latexml}
\iflatexml
  \usepackage{authblk}
\author[1]{Hyunwoo Yu}
\author[1]{Jongbeom Lee}
\author[1]{Hyunbin Kim}
\author[1]{Sungjun Park}
\author[1]{Jaemin Choi}
\author[1]{Jincheol Yang}
\author[1]{Seunghun Moon}
\author[2]{Bongjoon Hyun}
\author[2]{Kyomin Sohn}
\author[3]{Kyeongbo Kong}
\author[1]{Suk-Ju Kang}
\affil[1]{Sogang University}
\affil[2]{Samsung Electronics}
\affil[3]{Pusan National University}
\else
\author{\textbf{Hyunwoo Yu$^{1}$, Jongbeom Lee$^{1}$, Hyunbin Kim$^{1}$, Sungjun Park$^{1}$,} \\
\textbf{Jaemin Choi$^{1}$, Jincheol Yang$^{1}$, Seunghun Moon$^{1}$, Bongjoon Hyun$^{2}$,} \\
\textbf{Kyomin Sohn$^{2}$, Kyeongbo Kong$^{3}$, Suk-Ju Kang$^{1}$} \\
{\normalfont $^{1}$Sogang University \quad $^{2}$Samsung Electronics \quad $^{3}$Pusan National University} \\[4pt]
{\normalfont Project page: \url{https://hyunwoo137.github.io/OpenAttention/}} \\
{\normalfont Code: \url{https://github.com/hyunwoo137/OpenAttention}}}
\fi

\iclrfinalcopy
\hypersetup{hidelinks}
\begin{document}
\maketitle

% These links are ordinary content, not author affiliations, in HTML.
\iflatexml
\begin{center}
Project page: \url{https://hyunwoo137.github.io/OpenAttention/}\\
Code: \url{https://github.com/hyunwoo137/OpenAttention}
\end{center}
\fi
\lhead{Preprint}

\begin{abstract}
  Sparse attention reads only a small subset of the key-value cache, and how to treat the unread tail remains a key challenge. Discarding the tail biases the
  output, and sampling it as previous approaches do fails at reads of 5\% and below,
  where the tail likely holds heavy hitters the selector missed. Our key observation
  is that the simplest possible proxy, one token that folds the entire tail into its
  count-aware mean, already yields lower output error than a sample at every
  read behind a practical selector. We further show in closed form that at low reads
  sampling nearly doubles the weight error of discarding while a single proxy stays
  below it. Building on these findings, we explore approximating the dense attention
  distribution with a few proxies that target the tail. We propose OpenAttention, a pipeline of clustering, ErrorGate and top-k selection followed by a count-aware mean proxy.
  To implement clustering and selection both efficiently and effectively, we design training-free Query-Geometry Clustering (QGC) and a Query-Geometry Selector (QGS), which compress the keys while respecting the query directions. Additionally, unlike prior methods that rank clusters by centroid alone, ErrorGate allocates the exact read by the estimated weight error of each cluster.
  On RULER-32K HARD with Llama-3.1-8B, OpenAttention retains 99\% of dense
  accuracy at a 1\% read. It also achieves state-of-the-art accuracy across models, context lengths and datasets. Our
  customized attention kernel for OpenAttention runs the decode
  attention stage 10$\times$ faster than dense attention at a 1\% read.
\end{abstract}

\begin{figure}[!h]
  \centering
  \includegraphics[width=0.9\linewidth]{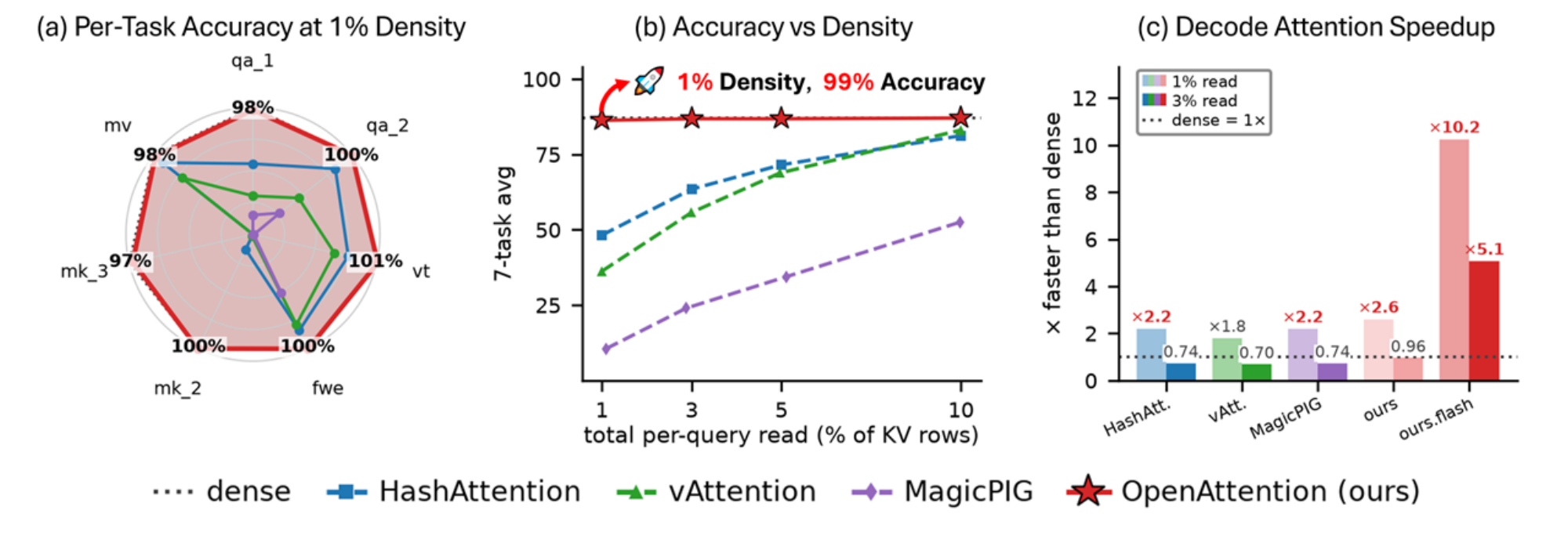}
  \vspace{-14pt}
  \caption{OpenAttention against sparse attention baselines on RULER-32K HARD with Llama-3.1-8B. (a) Per-task accuracy at a 1\% read. (b) Mean accuracy against the total per-query read. (c) Decode attention-stage speedup over dense attention at batch 32 and 32K context.}
  \label{fig:density}
\end{figure}

\section{Introduction}
\label{sec:intro}

\begin{figure}[t]
  \centering
  \includegraphics[width=0.96\linewidth]{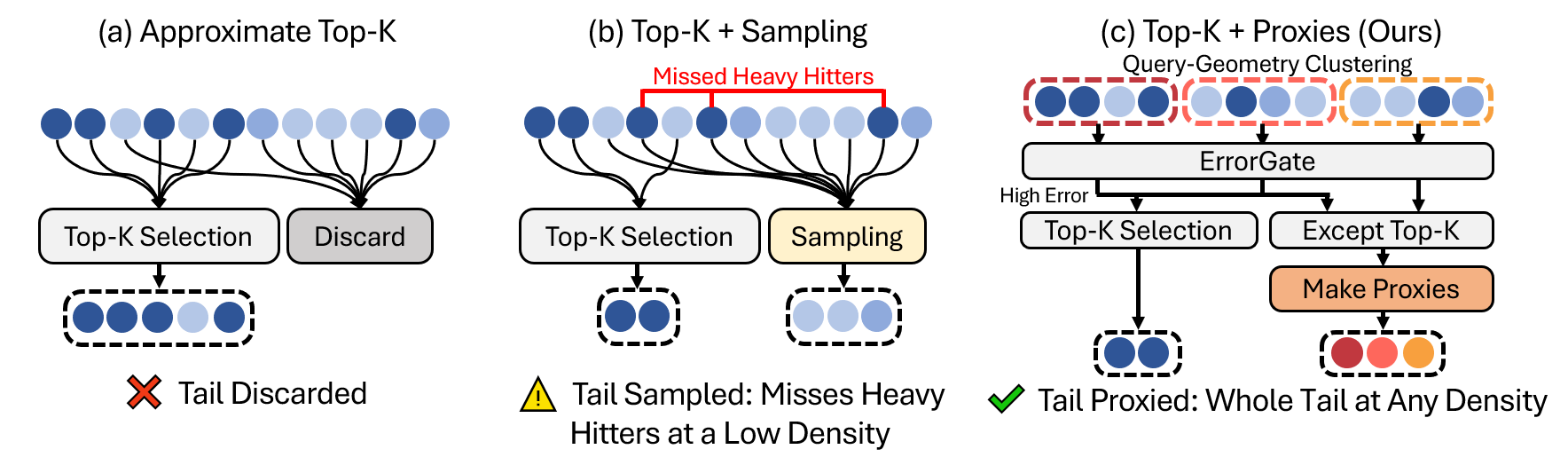}
  \vspace{-12pt}
  \caption{Three treatments of the tail for sparse attention. (a) Approximate top-k relies on top-k accuracy and ignores the tail. (b) Top-k with sampling misses heavy hitters at low density. (c) Ours clusters  and represents the tail by proxies built with a method that minimizes the weight error.}
  \label{fig:tailtreat}
  \vspace{-10pt}
\end{figure}

Long-context decoding is bound by the memory traffic of the key-value (KV) cache, and sparse attention reduces this cost by computing attention over only a subset of the cache. Existing top-k approaches \citep{singhania2024loki, yang2024doublesparsity, desai2024hashattention} explore frameworks that focus solely on predicting the top-k scores accurately and discard the remaining tail, as shown in Figure~\ref{fig:tailtreat}(a). However, omitting the tail information fundamentally prevents these methods from approximating the dense attention distribution. This limitation becomes especially critical in aggregation tasks (e.g., FWE) that depend on the entire distribution of attention weights. To address this problem, as shown in Figure~\ref{fig:tailtreat} (b), MagicPIG \citep{chen2024magicpig} estimates the output by locality-sensitive hashing (LSH) based importance sampling, and vAttention \citep{desai2026vattention} selects heavy hitters and then approximates the residual tail by a random sample sized to a relative error bound. Both methods presume that the softmax output of the tail can be estimated from a small sample. However, the estimation error of sampling grows sharply as the density decreases.

\begin{figure}[t]
  \centering
    \includegraphics[width=\linewidth]{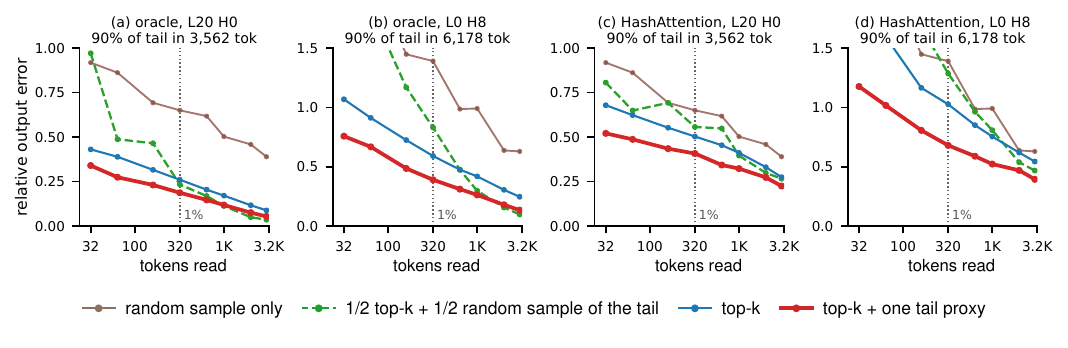}
  \vspace{-24pt}
  \caption{Relative attention-output error versus total tokens read for four tail treatments on two RULER-32K HARD attention heads, using oracle top-$k$ (a,b) or HashAttention (c,d).}
  \label{fig:tailcurves}
  \vspace{-10pt}
\end{figure}

Our key finding, as shown in Figure~\ref{fig:tailcurves}, is that a random sample complements the tail well at high density but yields a higher error than top-k alone at low density. However, folding the same tail into a single proxy with a simple count-aware mean keeps the error low in every setting we examine. This holds across different attention distributions and densities as well as behind both HashAttention and the oracle top-k selector. Based on this empirical finding, we measure the statistics of the tails left by HashAttention and the oracle top-k. Using these quantities, we show in closed form that a single proxy yields a lower attention weight error than sampling. Consequently, the tail should be approximated by well-constructed proxies rather than by samples, and this work explores how to build and exploit such proxies effectively.

As shown in Figure~\ref{fig:tailtreat}(c), we propose OpenAttention to approximate the tail distribution rather than truncate or sample the tail. OpenAttention constructs tail proxies through a framework that consists of clustering, ErrorGate, top-k selection and a count-aware mean. The effectiveness of this framework depends on how well it minimizes the weight error, which is governed by the mean and variance of each cluster. Two considerations are central to achieving this goal.
First, the clustering and the selector must run both effectively and efficiently to form clusters with small variance. Since this variance is measured in attention scores, the keys alone are insufficient and the query geometry must also be taken into account. We therefore compress the key tokens with the query eigenvectors and perform both clustering and top-k selection on the compressed keys. Second, clusters should be selected by their error, which then determines the top-k budget. Unlike prior methods that select clusters only by the mass of their centroid or mean, ErrorGate efficiently approximates the weight error from a mean token and an approximate variance token and uses it to allocate the budget.

% With these methods designed to maximize effectiveness within the proposed framework, OpenAttention preserves 99\% of dense accuracy at a 1\% read. On RULER-32K HARD~\citep{hsieh2024ruler} with Llama-3.1-8B~\citep{grattafiori2024llama3}, it scores 86.30 against 87.04 for full attention and 48.2 for the best baseline. OpenAttention also achieves state-of-the-art accuracy across different models (e.g., DeepSeek-R1-Distill-Llama-8B~\citep{deepseekai2025r1} and Mistral-7B-Instruct-v0.3~\citep{mistralai2024mistralv03}) and context lengths from 16K to 96K. Additionally, unlike prior sampling-based methods \citep{chen2024magicpig, desai2026vattention}, OpenAttention is designed to deliver practical latency gains. To this end, we customize a Triton-based FlashAttention-2~\citep{dao2023flashattention2} kernel to process the exact tokens and the proxies in an online softmax pass. The resulting kernel, OpenAttention.flash, runs the decode attention stage 10$\times$ faster than full attention at a 1\% read.

With these methods, OpenAttention preserves 99\% of dense accuracy at a 1\% read, scoring 86.30 against 87.04 for full attention and 48.2 for the best baseline on RULER-32K HARD~\citep{hsieh2024ruler} with Llama-3.1-8B~\citep{grattafiori2024llama3}. It also achieves state-of-the-art accuracy on DeepSeek-R1-Distill-Llama-8B~\citep{deepseekai2025r1} and Mistral-7B~\citep{mistralai2024mistralv03} and across context lengths from 16K to 96K. Moreover, unlike prior sampling-based methods~\citep{chen2024magicpig, desai2026vattention}, OpenAttention targets practical latency gains. Our Triton-based~\citep{tillet2019triton} customized kernel, OpenAttention.flash, processes the exact tokens and proxies in one online softmax pass and runs the decode attention stage 10$\times$ faster than dense attention at a 1\% read.

Our contributions are the following.
\begin{itemize}
  \item We propose \ours{}, which approximates the full softmax distribution rather than truncating or sampling the tail. It represents the tail by proxies built through clustering, \mgate{}, top-$k$ selection and a count-aware mean.
  \item We design \qgc{} and \qgs{}, which compress the key tokens under the query geometry so that clustering and top-$k$ selection follow the attention scores while remaining efficient.
  \item We introduce \mgate{}, which selects clusters and allocates the budget based on an approximated weight error rather than the mass of each cluster alone.
  \item \ours{} preserves 99\% of dense accuracy at a 1\% read on RULER-32K HARD with Llama-3.1-8B. It also achieves state-of-the-art accuracy across different models and context lengths from 16K to 96K.
  \item We customise a Triton-based FlashAttention-2~\citep{dao2023flashattention2} decode kernel into \oursflash{}, which runs the estimator as one online softmax pass. It accelerates the decode attention stage by $5$--$10\times$ over dense attention at 3\% and 1\% reads.
\end{itemize}
\section{Key observation on one simple proxy}
\label{sec:observation}
In this section, we examine how random sampling and a single simple proxy treat the tail. First, as shown in Figure~\ref{fig:tailcurves}, we analyze the empirical finding on the relative output error. Second, as shown in Figure~\ref{fig:formula}, we compare the closed-form weight errors of both treatments as functions of the tail statistics under a given top-$k$ selector.

\subsection{Relative output error across densities.}
As shown in Figure~\ref{fig:tailcurves}, we measure the relative output error of random sampling and our proxy at different densities. We evaluate two decode heads with different tail distributions, in which 90\% of the tail mass lies in 3,562 and 6,178 of the 32K tokens, respectively. Each head is evaluated behind two top-$k$ selectors, namely HashAttention and the oracle top-$k$. The sampling method splits the read equally between top-$k$ and a random sample of the tail, while the proxy method replaces the entire tail with a single proxy. The sampling method improves on its top-$k$ selector at high density but performs worse than the selector alone at low density. This indicates that a small sample cannot approximate the tail distribution as more tokens are left in the tail. In contrast, the single proxy reduces the error even in this simple form. Behind HashAttention it yields the lowest error at every density, and behind the oracle top-$k$ it does so at low density. Based on this empirical finding, we analyze the two treatments mathematically in the following subsection.

\subsection{Attention weights and tail approximation error}
\label{sec:theory}
\label{sec:setting}

\begin{figure}[t]
  \centering
  \includegraphics[width=0.9\linewidth]{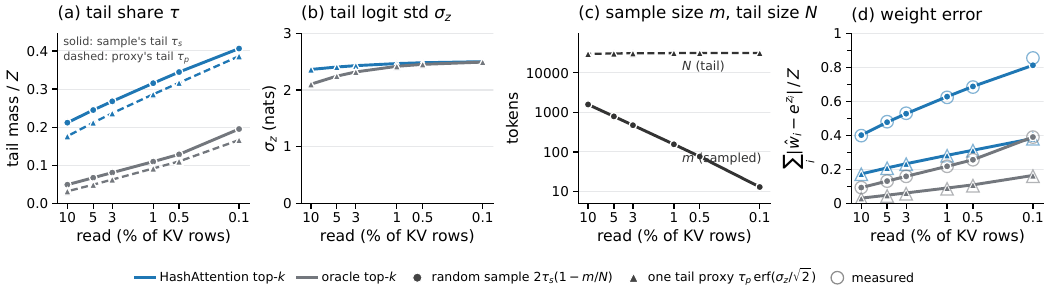}
  \vspace{-3pt}
  \caption{Weight-discrepancy comparison behind HashAttention and oracle
  top-$k$. Panels show the tail share, logit spread, sample/tail sizes,
  and the predicted and measured discrepancies of Eq.~(\ref{eq:tail-errors}).}
  \label{fig:formula}
  \vspace{-10pt}
\end{figure}

Let $w_i=e^{z_i}$ with $z_i=d^{-1/2}q^\top k_i$.
An exact set $X$ leaves tail $T=[n]\setminus X$, $N=|T|$, and
$Z_X=\sum_{i\in X}w_i$. For a uniform sample $B\subseteq T$ of size
$m$, let $\eta=N/m$ and $g_T=e^{\bar z_T}$. Following the deterministic
and randomized SDPA formulations of \citet{desai2026vattention},
discarding, sampling and one proxy assign
\begin{equation}
\hat a_i^{\mathrm d}
=\frac{\mathbf1_X(i)w_i}{Z_X},\qquad
\hat a_i^{\mathrm s}
=\frac{[\mathbf1_X(i)+\eta\mathbf1_B(i)]w_i}{\hat Z_{\mathrm s}},
\qquad
\hat a_i^{\mathrm p}
=\frac{\mathbf1_X(i)w_i+\mathbf1_T(i)g_T}{\hat Z_{\mathrm p}},
\label{eq:tail-weights}
\end{equation}
where
$\hat Z_{\mathrm s}=Z_X+\eta\sum_{j\in B}w_j$ and
$\hat Z_{\mathrm p}=Z_X+Ng_T$.
The proxy is equivalently one virtual token with logit
$\bar z_T+\log N$ and mean value $\bar v_T$.

For the implied unnormalised weights define
$\mathcal E=Z^{-1}\sum_i|\hat w_i-w_i|$,
$Z=\sum_iw_i$, and
$\tau=\sum_{i\in T}w_i/Z$. Then
\begin{equation}
\mathcal E_{\mathrm d}=\tau,\qquad
\mathbb E[\mathcal E_{\mathrm s}]
=2\tau\left(1-\frac mN\right),\qquad
\mathcal E_{\mathrm p}^{\mathrm G}
=\tau\,\mathrm{erf}\!\left(\frac{\sigma_z}{\sqrt2}\right).
\label{eq:tail-errors}
\end{equation}
The first two are exact for a fixed tail (the second in expectation);
the last is a Gaussian-population prediction. Under the fixed
unnormalised-weight convention above, sampling approaches twice the
discard discrepancy when $m\ll N$; this is not a factor-of-two claim
for normalized attention or output error.

Figure~\ref{fig:formula} shows, however, that the proxy and discard are
much closer in $\mathcal E$ than their output errors in
Figure~\ref{fig:tailcurves}. Thus $\mathcal E$ diagnoses unnormalised
weight distortion but does not by itself explain sampling's
density-dependent output failure. That dependence is captured by the
sampling-normalizer uncertainty, which grows as the sample shrinks;
missed high-weight tokens can also perturb the value direction
(App.~\ref{app:proxyformula}). The exact output identity in
App.~\ref{app:bound} explains why similar weight discrepancy need not
imply similar output error.

To connect this observation to the method, let $\mathcal P$ be a partition of the unread tokens into nonempty residual clusters, with sizes $n_b$ and mean logits $\bar z_b$. Their exact discrepancy decomposes as
\begin{equation}
\mathcal E=\frac1Z\sum_{b\in\mathcal P}\hat M_bD_b,\qquad
\hat M_b=n_be^{\bar z_b},\qquad
D_b=\frac1{n_b}\sum_{i\in b}|e^{z_i-\bar z_b}-1|.
\label{eq:identity}
\end{equation}
Each contribution $\varepsilon_b=\hat M_bD_b$ is proxy mass times weight spread; it vanishes for equal logits and is removed by reading the entire cluster exactly. Under the Gaussian model, $D_b$ is estimated by $D_{\mathrm G}(\sigma_b)\approx0.8\sigma_b$ at small within-cluster logit standard deviation $\sigma_b$. Writing $o=\sum_i a_iv_i$, $\hat o=\sum_i\hat a_iv_i$ and $\hat Z=\sum_i\hat w_i>0$, we also have $\|\hat o-o\|_2\le(Z/\hat Z)\mathcal E\max_i\|v_i-o\|_2$, without Gaussian or weight--value independence assumptions (App.~\ref{app:bound}). Section~\ref{sec:method} develops this multi-cluster construction using \qgc{} grouping, \mgate{} discrepancy-based allocation and \qgs{} selection within clusters.

\section{\ours{}: Approximate Tail Distribution by Proxies}
\label{sec:method}
% \begin{figure}[t]
%   \centering
%   \includegraphics[width=0.86\linewidth]{figs/fig_iclr_overall_method.png}
%   \vspace{-6pt}
%   \caption{Overview of \ours{}. (a) \qgc{} clusters the cache once per request. \mgate{} reads the top tokens of high-error clusters via the \qgs{}, and the rest of each cluster becomes one mass-preserving mean proxy. (b) \qgc{} clusters the keys projected onto the query eigenvectors. (c) \qgs{} scores the top-$k$ with low-bit projections of the keys and query.}
%   \label{fig:method}
%   \vspace{-10pt}
% \end{figure}

\subsection{Overall Process of OpenAttention}
\begin{figure}[t]
  \centering
  \includegraphics[width=0.86\linewidth]{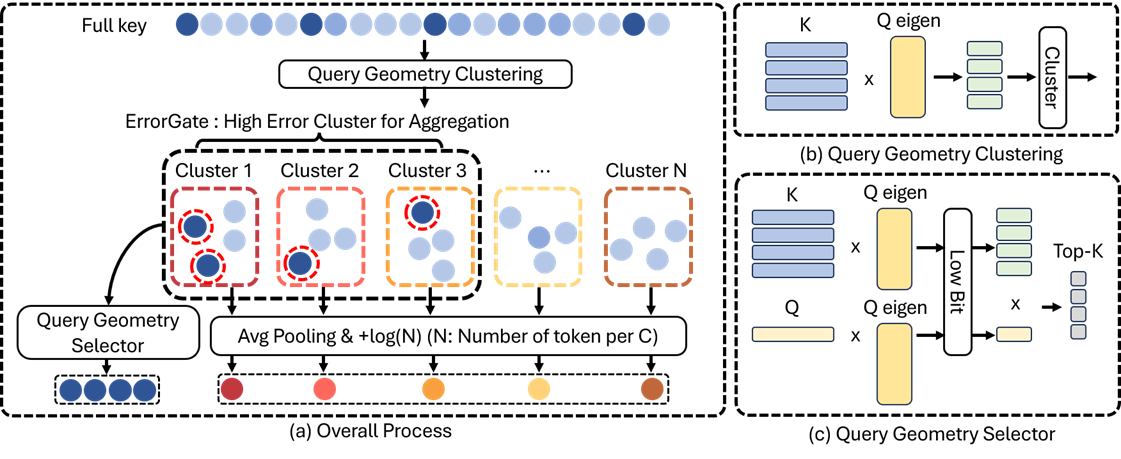}
  \vspace{-6pt}
  \caption{Overview of \ours{}. (a) \qgc{} clusters the cache once per request. \mgate{} reads the top tokens of high-error clusters via the \qgs{}, and the rest of each cluster becomes one count-aware mean logit proxy. (b) \qgc{} clusters the keys projected onto the query eigenvectors. (c) \qgs{} scores the top-$k$ with low-bit projections of the keys and query.}
  \label{fig:method}
  \vspace{-10pt}
\end{figure}

As shown in Figure~\ref{fig:method} (a), \ours{} clusters the cache once per request into $C$ clusters under the query geometry (\qgc{}). Each cluster caches its size $n_b$, a \emph{mean token} $(\bar k_b,\bar v_b)$ and a \emph{variance token} $\mathrm{diag}\,\mathrm{Var}_b(k)$. For each query, two small matrix products give every $\bar z_b$ and $\sigma_b^2$, from which \mgate{} estimates the error each proxy would leave. The exact budget $E$ is split in proportion to this estimate, and each cluster reads its highest-scoring tokens under the \qgs{}. The remaining members of each cluster then take the role of the tail $T$ in $\hat a^{\mathrm p}_i$ of Eq.~(\ref{eq:tail-weights}). Each cluster thus enters the softmax as one virtual token with the mean value and the logit $\bar z_b+\log n_b$ of its remaining members.
 The exact set $X$ now includes the initial sink, local-window and selector top-$k$ tokens as well as the $E$ allocated reads. The residual-proxy estimator in Eq.~(\ref{eq:merge}) merges all exact tokens and proxies in one online softmax. The initial exact set is small in the deployed 1\% configuration, and \mgate{} allocates the expansion budget rather than eliminating the initial selection stage.

%  As shown in Fig.~\ref{fig:method} (a), once per request, \ours{} clusters the cache into $C$ clusters under the query geometry (\qgc{}) and caches per cluster a \emph{mean token} $(\bar k_b,\bar v_b)$ with its size $n_b$ and a \emph{variance token} $\mathrm{diag}\,\mathrm{Var}_b(k)$. Per query, two small matrix products give every $\bar z_b$ and $\sigma_b^2$, from which \mgate{} estimates the error each proxy would leave. The exact budget $E$ is split in proportion to that estimate, each cluster reads its highest scoring tokens under the \qgs{}, and what stays enters as a proxy over the remaining members and is merged in one online softmax,
 %\begin{equation}
   %\hat o=\frac{\sum_{i\in X}e^{z_i}v_i+\sum_{b}e^{\ell_b}\bar v_b'}{\sum_{i\in X}e^{z_i}+\sum_{b}e^{\ell_b}},\qquad \ell_b=\bar z_b'+\log n_b',
   %\label{eq:merge}
 %\end{equation}
% where $X$ includes the initial sink, local-window and selector top-$k$ tokens as well as the $E$ allocated reads; primes denote statistics over the remaining members. The initial exact set is small in the deployed 1\% configuration, and \mgate{} allocates the expansion budget rather than eliminating the initial selection stage.

% \paragraph{\oursflash{}: \ours{} on FlashAttention-2.}
% \label{sec:kernel}
% Because a proxy has the same form as a token, we customise a Triton-based FlashAttention-2 decode kernel into \oursflash{} (Algorithm~\ref{alg:flash}), which processes the exact tokens and the proxies in one online softmax pass. Each program gathers, in tiles of 64, the selected tokens, the tokens read from expanded clusters and the $(\bar k_b,\bar v_b)$ of the clusters with $+\log n_b$ on the logit. Pooled sums are cached per cluster, so removing the tokens a query reads costs one subtraction per token the kernel already loads. The work per step is fixed, with no data dependent loop bound and no host synchronisation.

\subsection{Query geometry clustering and selection}
\begin{wrapfigure}[12]{r}{0.34\linewidth}
  \vspace{-12pt}
  \centering
  \includegraphics[width=\linewidth]{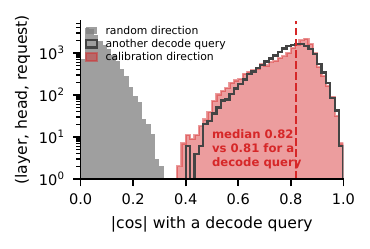}
  \vspace{-16pt}
  \caption{$|\cos|$ of a decode query with the leading eigenvector of $M$, another decode query and a random direction.}
  \label{fig:qsim}
  \vspace{-18pt}
\end{wrapfigure}

Both \qgc{} and \qgs{} use the top-$r$ eigenvectors of the queries to account for the query geometry when clustering and selecting key tokens. The queries used to obtain these eigenvectors combine pre-sampled query tokens ($\bar{M}_{\mathrm{st}}$) with the query tokens of the current question chunk ($\bar{M}_{q}$) as $M = \frac{1}{2}\bar{M}_{\mathrm{st}} + \frac{1}{2}\bar{M}_{q}$, and the top-$r$ eigenpairs $(U_r, \Lambda_r)$ of $M$ are used. These eigenpairs are computed once at the prefill stage. Clustering is then performed once after prefill, while selection reuses them at every decoding step. Therefore, as shown in Figure~\ref{fig:qsim}, we conduct an experiment to verify how well the $M$ obtained at prefill is preserved across decoding steps. The median absolute cosine similarity between a decode query and the leading eigenvector of $M$ is 0.82, which confirms that the query geometry is well preserved.

% \qgc{} and the \qgs{} need the same thing, a cheap prediction of which keys a query will score alike and which it will score highest, and the geometry of the queries provides both (Fig.~\ref{fig:qsim}). Per head, the queries concentrate in a few directions, and those directions persist from prefill to decoding: a decode query has a median $|\cos|$ of \QSIMCOSCAL{} with the leading eigenvector of $M=\mathbb{E}[qq^\top]$ estimated from the prefill queries of \emph{other} requests, as much as with a second decode query of the same head (\QSIMCOSSELF{}), against \QSIMCOSRAND{} for a random direction. The logits of a head therefore vary along the top eigenvectors of $M$, and projecting a key onto them keeps the part of it the queries see. In \ours{}, a static $M_{\text{st}}$ per layer and head is measured once on the queries of a small corpus outside the benchmark (18 LongBench~\citep{bai2024longbench} requests; before the rotary position embedding, RoPE, and rotated to the decode position) and shrunk toward the request's own question rows when a question chunk is available, $M=\tfrac12\bar M_{\text{st}}+\tfrac12\bar M_q$; its top-$r$ eigenpairs $(U_r,\Lambda_r)$ are all that \qgc{} and the \qgs{} use.

\subsubsection{\qgcfull{}}
\label{sec:partition}
Rather than grouping tokens by their keys alone, \qgc{} aims to group the tokens whose \emph{attention scores} will be close, taking the geometry of the queries into account. Since $\mathbb{E}_q(z_i-z_j)^2=s^2(k_i-k_j)^\top M(k_i-k_j)$, minimizing the within-cluster logit variance reduces to $k$-means under the metric $M$. Therefore, as shown in Figure~\ref{fig:method}(b), \qgc{} projects the keys onto the query eigenvectors and performs clustering in this projected space. With $K\in\mathbb R^{n\times d}$ the keys of a head and $(U_r,\Lambda_r)$ the top-$r$ eigenpairs of $M$, $r{=}16$,
\begin{equation}
  \Phi=K\,U_r\Lambda_r^{1/2}\in\mathbb R^{n\times r},\qquad
  a_i=\arg\min_b\,\|\phi_i-\mu_b\|^2,\qquad
  \mu_b=\frac1{n_b}\sum_{i:\,a_i=b}\phi_i ,
  \label{eq:qgc}
\end{equation}

\begin{wraptable}[9]{r}{0.45\linewidth}
\vspace{-14pt}
\centering
\caption{RULER-32K HARD mean accuracy by clustering method at a 1\% read ($C{=}64$).}
\label{tab:clustinput}
\setlength{\tabcolsep}{3pt}
\begin{tabular}{lcc}
\toprule
clustering & dim & avg \\
\midrule
token position & -- & 84.37 \\
$k$-means on keys only & 128 & 85.42 \\
\textbf{\qgc{} (ours)} & \textbf{16} & \textbf{86.30} \\
\bottomrule
\end{tabular}
\vspace{-10pt}
\end{wraptable}
\noindent where $\phi_i^\top$ denotes the $i$-th row of $\Phi$. Under this projection, the Euclidean distance between two projected keys matches their expected logit distance within the top-$r$ subspace. A few Lloyd iterations of the assignment and centroid steps in Eq.~(\ref{eq:qgc}) then yield the $C$ clusters. Each cluster then caches its count $n_b$, a mean token $(\bar k_b,\bar v_b)$ of its original keys and values and a variance token $\mathrm{diag}\,\mathrm{Var}_b(k)$ of per-dimension key variances.

\textbf{Validation of QGC.} To validate the role of query geometry, we compare \qgc{} with clustering by token position and with clustering on uncompressed keys alone~\citep{hu2025centroidkv,lu2025retroinfer}. As shown in Table~\ref{tab:clustinput}, \qgc{} is the most effective of the three while clustering in only 16 dimensions.

\subsubsection{\qgsfull{}}
\label{sec:selector}
\begin{wraptable}[13]{r}{0.5\linewidth}
  \vspace{-14pt}
  \centering
  \caption{Top-32 logit recall at $1/16$ and $1/32$ reads and selection time per layer on one H100.}
  \label{tab:selectors}
  \small
  \setlength{\tabcolsep}{4pt}
  \begin{tabular}{lccc}
    \toprule
    selector & $16\times$ & $32\times$ & ms \\
    \midrule
    SparQ & 0.96 & 0.94 & 0.29 \\
    Loki & 0.84 & 0.75 & 0.18 \\
    Quest & 0.77 & 0.66 & 0.33 \\
    Double Sparsity & 0.97 & 0.95 & 0.50 \\
    Squeezed Attention & 0.92 & 0.86 & 0.23 \\
    HashAttention & 0.74 & 0.63 & 0.24 \\
    \textbf{\qgs{} (ours)} & \textbf{0.97} & \textbf{0.94} & 0.22 \\
    \bottomrule
  \end{tabular}
  \vspace{-10pt}
\end{wraptable}
As shown in Figure~\ref{fig:method}(c), \qgs{} uses the eigenvectors of the same $M$ at rank $r{=}32$. Because selection runs at every decoding step, each key is coded once in low bits as its projection onto the query subspace to reduce latency. A query projected onto the same subspace then scores these codes as
\begin{equation}
  c_i=\mathrm{Q}_8\!\bigl(U_r^\top k_i\bigr),\qquad
  \tilde z_i=(U_r^\top q)^\top c_i ,
  \label{eq:qgs}
\end{equation}
where $\mathrm{Q}_8$ is int8 quantization with one scale per head and coordinate. The score $\tilde z_i\approx q^\top U_rU_r^\top k_i$ is the logit restricted to the query subspace, and its top-$k$ set $\mathcal T=\operatorname{top}\text{-}k(\tilde z)$ is read exactly. 
One $U$ is shared by all query heads of a KV head, and new keys are coded on arrival at $8r$ bits per token.

\textbf{Top-$k$ accuracy.}
We compare \qgs{} with six published top-$k$ selectors on the same decode rows using their released configurations. We keep the first and last 128 tokens at a $1/16$ read and 64 each at a $1/32$ read. The evaluation covers all layers and heads over the first eight decode steps of \SELNREQ{} samples. As shown in Table~\ref{tab:selectors}, \qgs{} achieves high top-$k$ recall at a low selection time, giving a favorable balance between accuracy and cost.

\subsection{\mgate{}}
\label{sec:rank}
\textbf{Estimating cluster discrepancy.}
Equation~(\ref{eq:identity}) gives the exact contribution $\varepsilon_b=\hat M_bD_b$, $\hat M_b=n_be^{\bar z_b}$. A Gaussian model replaces $D_b$ by $D_{\mathrm G}(\sigma_b)=e^{\sigma_b^2/2}(2\Phi(\sigma_b)-1)$; diagonal key moments approximate the exact variance $\sigma_b^2=s^2q^\top\mathrm{Cov}_b(k)q$ (Corollary~\ref{cor:jgap}). The deployed score is
\begin{equation}
  \begin{aligned}
    \hat\varepsilon_b&=n_be^{\bar z_b}\,e^{\hat\sigma_b^2/2}\bigl(2\Phi(\hat\sigma_b)-1\bigr)
    =\hat M_bD_{\mathrm G}(\hat\sigma_b),\\
    \bar z_b&=s\,q^\top\bar k_b,\qquad
    \hat\sigma_b^2=s^2\textstyle\sum_dq_d^2\,\mathrm{Var}_b(k_d).
  \end{aligned}
  \label{eq:ucb}
\end{equation}

\begin{wraptable}[9]{r}{0.4\linewidth}
  \vspace{-6pt}
  \centering
  \caption{\mgate{} ranking ablation at a 1\% read on RULER-32K HARD.}
  \label{tab:rankrule}
  \setlength{\tabcolsep}{4pt}
  \begin{tabular}{lc}
    \toprule
    ranking statistic & avg \\
    \midrule
    \tabrankrulerows
    \bottomrule
  \end{tabular}
  \vspace{-10pt}
\end{wraptable}
\noindent It uses cached mean and diagonal variance tokens and one error function per cluster. The signed mean enters through $e^{\bar z_b}$, and $D_{\mathrm G}(\hat\sigma_b)\approx0.8\hat\sigma_b$ at small dispersion. Thus $\sum_b\hat\varepsilon_b$ estimates $Z\mathcal E$ rather than equalling it. The score is neither an exact finite-cluster discrepancy nor a certified output bound; no independence assumption between weights and values is used. The signed mass factor is retained: for fixed cluster size and dispersion, $e^{\bar z_b}$ distinguishes a cluster the query aligns with from one it looks away from, whereas a statistic based on $\bar z_b^2$ does not. 
As shown in Table~\ref{tab:rankrule}, ranking clusters by the estimated error of Eq.~(\ref{eq:ucb}) yields 86.30 at a 1\% read, whereas ranking by the centroid mass as in prior methods~\citep{lu2025retroinfer, hu2025centroidkv} yields 85.99. This shows that accounting for within-cluster dispersion selects clusters more effectively than centroid mass alone.

\begin{wrapfigure}[11]{r}{0.3\linewidth}
  \vspace{-16pt}
  \centering
  \includegraphics[width=\linewidth]{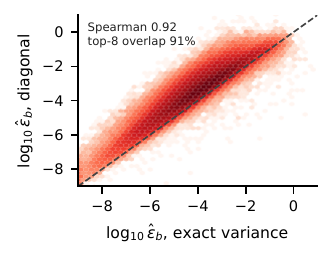}
  \vspace{-20pt}
  \caption{\mgate{} estimates from diagonal and full within-cluster logit variance on \CDNROWS{} decode rows.}
  \label{fig:errgate}
  \vspace{-12pt}
\end{wrapfigure}

\textbf{Validity of the diagonal estimate.}
The mean and variance tokens cost 1.5 KV rows per cluster per query, charged to the read. On the $C{=}64$ QGC diagnostic, diagonal and full-variance model scores agree at Spearman \ESTRHODF{} (Fig.~\ref{fig:errgate}) and recover the eight largest
exact-discrepancy clusters at similar rates (\GATEOVDT{}, \GATEOVFT{}). These measurements assess the diagonal substitution empirically; they are not a general variance inequality.

\textbf{Hierarchical budget.}
For a fixed number of complete clusters, removing the largest exact $\varepsilon_b$ terms maximally reduces $\mathcal E$. Deployment instead uses estimated scores and a token budget. Motivated by the measured concentration of discrepancy (App.~\ref{app:errgate}), we allocate
\begin{equation}
  q_b=\min\!\left(n_b,\lambda\frac{\hat\varepsilon_b}{\max_{b'}\hat\varepsilon_{b'}}\right),\qquad \sum_bq_b=E,
  \label{eq:quota}
\end{equation}
with $\lambda$ set by water filling. Leftover tokens after flooring go to the highest-ranked clusters with capacity so that the budget $E$ is spent exactly. Each cluster then reads its $q_b$ highest \qgs{}-scored tokens and recomputes its residual proxy statistics. This proportional allocation is a heuristic rather than an optimal rule. Scores estimate discrepancy before allocation; Proposition~\ref{prop:bound} applies after residual recomputation. Allocation details are in App.~\ref{app:errgate}.

\subsection{\oursflash{}: \ours{} on FlashAttention-2.}
\label{sec:kernel}
Because a proxy has the same form as a token, we customise a Triton-based FlashAttention-2 decode kernel into \oursflash{} (Algorithm~\ref{alg:flash} in App.~\ref{app:kernel}), which processes the exact tokens and the proxies in one online softmax pass. Each program gathers, in tiles of 64, the selected tokens, the tokens read from expanded clusters and the $(\bar k_b,\bar v_b)$ of the clusters with $+\log n_b$ on the logit. Pooled sums are cached per cluster, so removing the tokens a query reads costs one subtraction per token the kernel already loads. The work per step is fixed, with no data dependent loop bound and no host synchronisation.

\section{Experiments}
\label{sec:results}

\paragraph{Setup.} We use Llama-3.1-8B-Instruct (bf16, 32 query and 8 KV heads) on RULER-32K HARD, the seven tasks that HashAttention and \vatt{} identify as challenging for sparse attention (qa\_1, qa\_2, vt, fwe, niah\_multikey\_2/3, niah\_multivalue), with 200 samples per task. Following HashAttention and \vatt{}, the context is prefilled densely, sparse attention applies to the question chunk and generation, and the baselines share a sink and a local window of 0.1\% each. The baselines run in one framework on the HashAttention selector with the public signature weights, \vatt{} under its authors' per budget protocol and \mpig{} with $(K,L)$ realising the target read (App.~\ref{app:fairsplit}, \ref{app:baselines}). \ours{} uses the \qgs{}, \qgc{}, and Eq.~(\ref{eq:ucb}) with the hierarchical budget. Sink and local remain at their few-token floor (0.01\% each). \emph{Budget accounting.} We charge every stream a query reads in KV rows per query head, including the mean and variance tokens of all $C$ clusters. The index scan is excluded from the budget for every method. The transfer experiments on other models, datasets and context lengths use the conventional 3\% split (App.~\ref{app:fairsplit}).

\subsection{Comparison with Other Methods across Read Densities}
\begin{figure}[t]
\centering
\includegraphics[width=\linewidth]{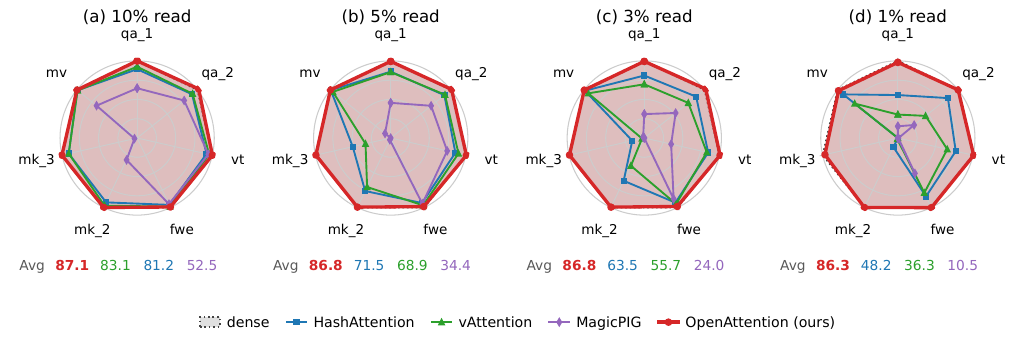}
\vspace{-18pt}
\caption{
RULER-32K HARD with Llama-3.1-8B-Instruct at matched total per-query reads of 10\%, 5\%, 3\%, and 1\%, including the proxy stream.
Each panel shows per-task accuracy relative to dense (grey, 100\%), with mean accuracy below.
Dense has a mean of 87.04 on RULER hard.
\mpig{} realises reads of 10.0\%, 5.1\%, 2.9\%, and 1.1\%.
Per-task results are in App.~\ref{app:densityfull}.
}
\label{fig:taskradars}
\vspace{-10pt}
\end{figure}

As shown in Figure~\ref{fig:taskradars}, we compare per-task accuracy on RULER-32K HARD at densities from 10\% to 1\%. HashAttention discards the tail while \mpig{} and \vatt{} rely on sampling, and all three degrade sharply as the density decreases. \vatt{} slightly improves on HashAttention at a 10\% density but falls below it at lower densities even though both share the same selector. In contrast, \ours{} preserves accuracy close to dense attention at every density. This demonstrates that approximating the tail with proxies is more effective than sampling it.

\begin{table}[t]
  \centering
  \begin{minipage}[t]{0.40\linewidth}
    \centering
    \caption{Ablation at a 5\% read on RULER-32K HARD (dense 87.04).}
    \label{tab:ablation}
    \small
    \setlength{\tabcolsep}{4pt}
    \begin{tabular}{cccc}
      \toprule
      proxies & \mgate{} & \qgs{} & avg \\
      \midrule
      \xmark & \xmark & \xmark & 67.35 \\
      \cmark & \xmark & \xmark & 77.91 \\
      \cmark & \cmark & \xmark & 85.90 \\
      \cmark & \xmark & \cmark & 86.73 \\
      \cmark & \cmark & \cmark & \textbf{86.82} \\
      \bottomrule
    \end{tabular}
  \end{minipage}\hfill
  \begin{minipage}[t]{0.57\linewidth}
    \centering
    \caption{Mean accuracy on RULER-32K HARD at a 3\% read on three models.}
    \label{tab:multimodel}
    \small
    \setlength{\tabcolsep}{4pt}
    \renewcommand{\arraystretch}{1.22}
    \begin{tabular}{lccc}
      \toprule
      method & Llama-8B & DeepSeek-8B & Mistral-7B \\
      \midrule
      dense & 87.04 & 73.46 & 65.46 \\
      HashAttention & 63.48 & 59.85 & 51.24 \\
      \vatt{} & 57.15 & 49.26 & 46.14 \\
      \textbf{\ours{}} & \textbf{86.59} & \textbf{73.49} & \textbf{64.83} \\
      \bottomrule
    \end{tabular}
  \end{minipage}
\end{table}
\subsection{Ablation Study}

Table~\ref{tab:ablation} compares the effects of proxies, \mgate{}, and \qgs{} under the same 5\% total-read budget. The configuration without any of the three components corresponds to \vatt{} on the HashAttention selector. With HashAttention as the selector, using proxies instead of residual sampling raises mean accuracy on RULER-32K HARD from 67.35 to 77.91. Adding \mgate{} alone to the proxy variant yields 85.90, while adding \qgs{} alone yields 86.73. Enabling both yields 86.82. Table~\ref{tab:rankrule} holds the deployed 1\% configuration fixed and varies only the \mgate{} ranking statistic. Ranking by the estimated error of Eq.~(\ref{eq:ucb}) yields 86.30 mean accuracy on RULER-32K HARD, compared with 85.99 for mean logit and 84.81 for variance alone.

\subsection{Evaluation on various models}
\label{sec:multimodel}

To evaluate \ours{} on models other than Llama-3.1-8B-Instruct, we run the same comparison on DeepSeek-R1-Distill-Llama-8B and Mistral-7B-Instruct-v0.3. All methods are evaluated at a 3\% density. As shown in Table~\ref{tab:multimodel}, \ours{} stays within 0.7 points of dense on all three models while the baselines lose up to 29.9 points. These results show that \ours{} effectively preserves accuracy across different models.

\subsection{Orthogonality to KV Cache Compression}
\label{sec:kvcompress}

Sparse attention is closely related to KV cache compression but can be applied orthogonally to it. We therefore apply \ours{} on top of three representative compressors, PyramidKV~\citep{cai2024pyramidkv}, SnapKV~\citep{li2024snapkv} and StreamingLLM~\citep{xiao2023streamingllm}. \ours{} then reads 10\% of the original context from the entries each compressor retains. As shown in Table~\ref{tab:kvcompress}, adding \ours{} changes the mean accuracy by only \KVCDELTAAVG{} points on average. KV cache compression saves memory by permanently evicting keys but loses accuracy under a small budget, whereas sparse attention keeps every entry accessible and preserves accuracy.

\subsection{Context length}
\label{sec:length}
Table~\ref{tab:length} compares mean accuracy at a matched 3\% total read as the context grows from 16K to 96K, using 100 samples per task. Dense accuracy falls from 87.14 to 78.01 over this range while \ours{} changes from 87.49 to 78.10. The difference between them remains below 0.7 points at every length, and \ours{} is numerically higher at 16K and 96K. At 96K HashAttention scores 27.99 while \vatt{} scores 40.20 and \mpig{} scores 33.19. These results show that \ours{} maintains accuracy close to dense attention across context lengths even as the other sparse methods fall substantially behind. Per-task scores and actual read rates are reported in App.~\ref{app:length}.

% Table~\ref{tab:length} compares mean accuracy at a matched 3\% total read across 16K, 32K, 64K, and 96K contexts. Dense accuracy falls from 87.14 to 78.01 over this range while \ours{} changes from 87.49 to 78.10. The difference between them remains below 0.7 points at 16K, 32K, 64K, and 96K, with \ours{} numerically higher at 16K and 96K. At 96K HashAttention scores 27.99 while \vatt{} scores 40.20 and \mpig{} scores 33.19. These results show that \ours{} maintains accuracy close to dense attention across context lengths even as the other sparse methods fall substantially behind. Per task scores and actual read rates are reported in App.~\ref{app:length}.

% \subsection{Latency}
% \label{sec:latency}
% Figure~\ref{fig:density}(c) and Table~\ref{tab:latency} (App.~\ref{app:kernel}) isolate the decode attention step of one layer on an H100 NVL (Llama-3.1-8B shapes, bf16). At a 3\% read, batch 32 and 32K context, \oursflash{} takes 0.24\,ms against 1.22\,ms for fused dense scaled dot product attention (SDPA), while the unfused gather paths of the baselines are \emph{slower} than dense (1.65--1.75\,ms). At 1\% it takes 0.12\,ms, $10\times$ below dense, while the baselines stay $4$--$6\times$ above it. The asymmetry is structural, since \vatt{}'s data dependent budget needs a host round trip per layer and \mpig{} runs this stage on the CPU. These timings use $E{=}0$. The deployed split takes 0.114\,ms at 1\% and 0.233\,ms at 3\%. Selection and \mgate{} scoring are excluded.

\begin{table}[t]
  \centering
  \begin{minipage}[t]{0.42\linewidth}
    \centering
    \caption{KV cache compression with sparse reading on RULER-32K HARD. Each compressor retains \KVCBUDGET{} entries.}
    \label{tab:kvcompress}
    \small
    \setlength{\tabcolsep}{4pt}
    \renewcommand{\arraystretch}{1.22}
    \begin{tabular}{lcc}
      \toprule
      compressor & alone & $+$\,\ours{} \\
      \midrule
      dense attention & 87.04 & -- \\
      StreamingLLM & \KVCALONEstreamingllm{} & \KVCPAIRstreamingllm{} \\
      SnapKV & \KVCALONEsnapkv{} & \KVCPAIRsnapkv{} \\
      PyramidKV & \KVCALONEpyramidkv{} & \KVCPAIRpyramidkv{} \\
      \bottomrule
    \end{tabular}
  \end{minipage}\hfill
  \begin{minipage}[t]{0.55\linewidth}
    \centering
    \caption{Mean accuracy on RULER-32K HARD at a 3\% read as the context grows. \ours{} scales only $C$ with the length.}
    \label{tab:length}
    \small
    \setlength{\tabcolsep}{4pt}
    \begin{tabular}{lcccc}
      \toprule
      method & 16K & 32K & 64K & 96K \\
      \midrule
      \tablenavg
      \bottomrule
    \end{tabular}
  \end{minipage}
\end{table}

\section{Related work}
\label{sec:related}

% \paragraph{Top $k$ selectors.}
\paragraph{Top-$k$ selectors.}
Quest~\citep{tang2024quest} and InfLLM~\citep{xiao2024infllm} score tokens at a coarse granularity through page-level bounds and block representatives. Learning-to-hash~\citep{sun2022learningtohash} and HashAttention~\citep{desai2024hashattention} are hashing-based methods that map queries and keys to compact codes and retrieve tokens by code similarity. Loki~\citep{singhania2024loki} and Squeezed Attention~\citep{hooper2024squeezed} rely on the keys alone and score tokens in a PCA subspace of the keys or through offline key clusters. PQCache~\citep{zhang2024pqcache} applies product quantization to the keys and retrieves tokens by approximate inner product search. In contrast, RetrievalAttention~\citep{liu2024retrievalattention}, SparQ~\citep{ribar2024sparq} and Double Sparsity~\citep{yang2024doublesparsity} exploit query information to process the keys. RetrievalAttention builds a key index that accounts for the query distribution, while SparQ and Double Sparsity score the keys on channels chosen from the query at each step or by offline calibration. \qgs{} follows this line of work. However, it differs in that it projects both the keys and the query onto the leading eigenvectors of the query second moment $M$. The same eigenvectors also define the metric of \qgc{}, so selection and clustering share one query geometry. Although accurate top-$k$ prediction is essential, top-$k$ selection alone cannot account for the distribution of the tail tokens.

% Sparse attention reads the rows that a cheap score predicts to matter at each step.
% Existing methods use page level bounds~\citep{tang2024quest} or block and chunk representatives~\citep{xiao2024infllm,sun2024shadowkv}.
% Others use nearest neighbour indices over the keys~\citep{liu2024retrievalattention,zhang2024pqcache,mao2024iceformer} or learned signatures~\citep{sun2022learningtohash,desai2024hashattention}.
% Double Sparsity~\citep{yang2024doublesparsity} selects $r$ coordinate channels using an offline outlier statistic.
% SparQ~\citep{ribar2024sparq} selects them per query from the largest components of $q$.
% Loki~\citep{singhania2024loki} scores in a low rank subspace derived from the keys while the \qgs{} derives its subspace from the queries.
% Table~\ref{tab:selectors} compares these methods at the same rank and read.
% NSA~\citep{yuan2025nsa} includes a compressed token branch and SALS~\citep{mu2025sals} performs selection in a latent space.
% Selecting high scoring tokens alone leaves the contribution of unread tokens unresolved.

% \paragraph{Sampling the tail.}
\paragraph{Sampling the tail.}
Rather than discarding the tail, \mpig{}~\citep{chen2024magicpig} estimates the attention output by importance sampling, drawing keys through LSH collisions and weighting them by their collision probabilities. \vatt{}~\citep{desai2026vattention} first selects the top-$k$ tokens and then approximates the remaining tail with a random sample. Its sample size is set from a central limit theorem bound so that the estimate meets a user-specified error tolerance. SANTA~\citep{lee2026santa} samples value rows after the softmax. However, at a low read these samples can miss the heavy hitters a practical selector leaves in the tail, so we instead summarize every tail token with proxies. SparQ~\citep{ribar2024sparq} also folds the unread tokens into one mean value vector but weights it by the leftover approximate score mass rather than a count-aware mean logit.

% Rather than discarding the tail, \mpig{}~\citep{chen2024magicpig} estimates the output using LSH importance sampling on the CPU.
% \vatt{}~\citep{desai2026vattention} adds random residual sampling whose size is determined by a central limit theorem criterion.
% SANTA~\citep{lee2026santa} samples value rows after the softmax.
% At a fixed low read residual samples can miss heavy hitters left behind by a practical selector.
% Under our weight discrepancy metric uniform residual sampling without replacement incurs $2\tau(1-m/N)$ in expectation as shown in Eq.~(\ref{eq:tail-errors}) 
% When $m\ll N$ this approaches twice the discrepancy of discarding as shown in Figure~\ref{fig:formula}.
% This result does not imply a universal ordering of output error across these estimators.
% SampleAttention~\citep{zhu2025sampleattention} identifies structured sparse patterns rather than estimating the residual tail output.

% \paragraph{Clustering and centroid methods.}
\paragraph{Clustering and centroid proxies.}
Clustering has been used for both KV cache compression and sparse attention. For compression, CentroidKV~\citep{hu2025centroidkv} merges each cluster of similar keys into a single centroid, while ClusterKV~\citep{liu2024clusterkv} keeps every token and recalls whole clusters by the attention weight of their centroids. For sparse attention, Squeezed Attention~\citep{hooper2024squeezed} clusters the keys of a fixed context offline and loads the keys whose centroids match the query. Double-P~\citep{doublep2026} estimates top-$p$ attention mass at the cluster level from size-weighted centroids and refines it at the token level only when needed. However, these methods cluster by key similarity alone, which limits how far they can reduce the variance of attention scores within each cluster. In addition, their cluster selection relies on centroid mass, which ignores the error of representing a cluster by its centroid. In contrast, \qgc{} clusters the keys under the query geometry, and \mgate{} selects clusters and allocates the budget by the estimated weight error.

\section{Conclusion and Limitations}
\label{sec:conclusion}
We propose \ours{}, which approximates the information of the tail with proxies to address the loss of tail information in sparse attention. \ours{} approximates the tail logits effectively through a framework of clustering, \mgate{}, top-$k$ selection and a count-aware mean. To optimize each stage, it exploits the query geometry and minimizes the estimated weight error. As a result, \ours{} preserves 99\% of dense accuracy at a 1\% read and achieves state-of-the-art accuracy across benchmarks and models.

Despite its effectiveness, \mgate{} approximates the within-cluster variance with its diagonal terms for efficiency. This approximation agrees closely with the full-variance estimate at a Spearman correlation of 0.92 but cannot recover the exact error. We believe that reducing this proxy error within the \ours{} framework is a promising direction for future work.

% \section{Limitations and conclusion}
% \label{sec:limitations}
% Reasoning or code heavy suites are not run, the accounting charges neither the index scan shared by all sparse methods (App.~\ref{app:kernel}) nor the per request build, and the clusters' decay under long generation is unmeasured. A sampled tail cannot represent the heavy hitters a selector misses at low density; \ours{} represents that tail by proxies over clusters the query geometry defines, lets the error of each proxy decide the read, and stays within one point of dense at a 1\% read.

\section*{AI use statement}
In this work, we used generative AI tools to edit the text of the paper for clarity and readability, to translate parts of the authors' draft into English, to help search for and identify related literature, to help write and format its \LaTeX{} source, including tables, captions and references, and to review the manuscript for consistency between the numbers reported in the text, tables and appendices. We have not used generative AI tools to propose or revise research hypotheses, to develop the method, its theoretical analysis or conceptual framework, to formulate the mathematical claims, to provide key ingredients of their proofs or to write the proofs, to design the experiments or provide feedback on the methodology or experiments, to implement the method or the baselines, to generate, clean or reformat datasets, including the synthetic 64K and 96K RULER sets, or to interpret the results. Qualitative or thematic analysis is not applicable to this work. The authors checked every AI-suggested edit, including translations, against our experiment logs and source code, and read every AI-suggested reference before citing it. We take responsibility for the final content of this work, including text, claims or artifacts produced with the aid of generative AI.

\section*{Ethics statement}
This work does not involve human subjects or personal data. All models and benchmarks we use are publicly available and are used under their licenses, and the 64K and 96K RULER sets we generate are synthetic. Like any approximate attention method, ours can change model outputs, so its accuracy should be validated on the target task before deployment.

\section*{Reproducibility statement}
The method is defined in \S\ref{sec:method}. QGC is given in Eq.~(\ref{eq:qgc}), the QGS in Eq.~(\ref{eq:qgs}), the \mgate{} score in Eq.~(\ref{eq:ucb}) and the budget allocation in Eq.~(\ref{eq:quota}). The read accounting and the evaluation protocol are given in \S\ref{sec:results}. The proofs and their assumptions are in App.~\ref{app:bound}, \ref{app:errgate} and~\ref{app:proxyformula}. The deployed hyper-parameters and the per-budget grid are in App.~\ref{app:fairsplit}, and per-task results at every budget are in App.~\ref{app:densityfull}. The baseline configurations are in App.~\ref{app:baselines}, \ref{app:magicpig} and~\ref{app:vattprotocol}. The kernel is described in App.~\ref{app:kernel} and Algorithm~\ref{alg:flash}, and the generation of the 64K and 96K RULER sets in App.~\ref{app:length}. All models and benchmarks are publicly available, and all accuracy numbers use greedy decoding with fixed prompts. Code is available at \url{https://github.com/hyunwoo137/OpenAttention}.

\bibliography{main}
\bibliographystyle{iclr2027_conference}

\appendix

\section{Error bound of cluster aggregation and the ranking rule}
\label{app:bound}
This appendix supplies the detailed setting and derivations for \S\ref{sec:theory}, and analyzes the residual-cluster estimator of \S\ref{sec:method}. Exact finite-cache identities are distinguished from Gaussian population estimates and allocation heuristics.

\paragraph{Detailed setting and tail estimators.}
For $q\in\mathbb R^d$ and $(K,V)\in\mathbb R^{n\times d}$, define $z_i=sq^\top k_i$, $s=d^{-1/2}$, $w_i=e^{z_i}$, $Z=\sum_iw_i$, $a_i=w_i/Z$ and $o=\sum_i a_iv_i$. 
A nonempty exact set $X$ leaves $T=[n]\setminus X$, with $Z_X=\sum_{i\in X}w_i$ and $S_T=\sum_{i\in T}w_i$. Writing $\mu_X=Z_X^{-1}\sum_{i\in X}w_iv_i$, a tail estimator has the form
\[
\hat o=\frac{Z_X\mu_X+\hat S\hat\mu}{Z_X+\hat S},
\]
where $\hat S$ and $\hat\mu$ represent the estimated tail mass and value mean. Discarding uses $\hat S=0$; uniform residual sampling uses the inverse-probability weights of Eq.~(\ref{eq:tail-weights}); and one proxy uses $\hat S=Ne^{\bar z_T}$ and $\hat\mu=\bar v_T$, where $N=|T|$ and both means are arithmetic means over $T$. Its mass equals $S_T$ only when the tail logits are equal. For the implied unnormalised token weights $\hat w_i$, define $\hat Z=\sum_i\hat w_i$, $\hat a_i=\hat w_i/\hat Z$ and $\mathcal E=Z^{-1}\sum_i|\hat w_i-w_i|$. The normalization of the unnormalised weights is fixed by the numerators in Eq.~(\ref{eq:tail-weights}); $\mathcal E$ is not $\|\hat a-a\|_1$. Accordingly, $\mathcal E$ is tied to this numerator convention and is not invariant to a common rescaling of the implied weights, whereas $\hat a$ is.

\paragraph{Residual proxy estimator.}
For the multi-cluster estimator, $X$ includes sink, local-window, selector top-$k$ and allocated exact reads. Let $\mathcal P$ contain only the nonempty residual clusters. For $b\in\mathcal P$, $n_b$, $\bar k_b$ and $\bar v_b$ are the count and arithmetic means \emph{after} exact-read tokens are removed. Set $\bar z_b=sq^\top\bar k_b$ and
$\hat M_b=n_be^{\bar z_b}$. The residual-proxy estimator is
\begin{equation}
\hat o=
\frac{
\sum_{i\in X}w_iv_i+
\sum_{b\in\mathcal P}\hat M_b\bar v_b
}{
\sum_{i\in X}w_i+
\sum_{b\in\mathcal P}\hat M_b
},
\qquad
\hat M_b=n_be^{\bar z_b}.
\label{eq:merge}
\end{equation}

Write $\delta_i=z_i-\bar z_b$ for $i\in b$, so $\sum_{i\in b}\delta_i=0$, and define
\[
\sigma_b^2=\frac1{n_b}\sum_{i\in b}\delta_i^2
=s^2q^\top\mathrm{Cov}_b(k)q,\qquad
\hat a_b^{\mathrm{virt}}=\frac{\hat M_b}{\hat Z},\qquad
V_b=\max_{i\in b}\|v_i-o\|_2.
\]
Also let $\Delta_b^+=\max_{i\in b}\delta_i$ and $\Delta_b^-=-\min_{i\in b}\delta_i$. The diagonal estimate in Eq.~(\ref{eq:ucb}) approximates $\sigma_b^2$. The virtual-token weight $\hat a_b^{\mathrm{virt}}$ is the sum of the implied original-token weights $\hat a_i=e^{\bar z_b}/\hat Z$ for $i\in b$.

\paragraph{Exact clusterwise discrepancy.}
Exact-read tokens have $\hat w_i=w_i$ and contribute no discrepancy; for $i\in b$, the proxy implies $\hat w_i=e^{\bar z_b}$. Factoring $e^{\bar z_b}$ within each cluster gives
\[
\mathcal E=\frac1Z\sum_{b\in\mathcal P}\hat M_bD_b,\qquad
D_b=\frac1{n_b}\sum_{i\in b}|e^{\delta_i}-1|,\qquad
\varepsilon_b=\hat M_bD_b,
\]
which proves Eq.~(\ref{eq:identity}).
This is exact for any finite residual partition, without a Gaussian assumption. Each term is nonnegative and vanishes for equal logits. Reading a whole cluster exactly removes its term. Partial reads change the residual mean and require recomputing the discrepancy.

\begin{lemma}[exact error of the estimator]\label{lem:identity}
  \begin{equation}
    \hat Z(\hat o-o)=-\sum_{b\in\mathcal P}\sum_{i\in b}
    \bigl(e^{z_i}-e^{\bar z_b}\bigr)(v_i-o).
    \label{eq:lemma-identity}
  \end{equation}
\end{lemma}
\begin{proof}
  Let $N_X=\sum_{i\in X}e^{z_i}v_i$ and $Z_X=\sum_{i\in X}e^{z_i}$. Then $\hat Z(\hat o-o)=N_X-oZ_X+\sum_b\hat M_b(\bar v_b-o)$. Since $o$ is the exact output, $\sum_{\text{all }i}e^{z_i}(v_i-o)=0$, hence $N_X-oZ_X=-\sum_b\sum_{i\in b}e^{z_i}(v_i-o)$. Finally $\hat M_b(\bar v_b-o)=\sum_{i\in b}e^{\bar z_b}(v_i-o)$, and the two sums combine into Eq.~(\ref{eq:lemma-identity}).
\end{proof}

\paragraph{Relation to normalized attention and output.}
The normalized-attention bridge of \S\ref{sec:setting} follows without a model assumption. For any nonnegative implied weights with $\hat Z>0$, set $\tilde a_i=\hat w_i/Z=(\hat Z/Z)\hat a_i$. Then $\|\tilde a-a\|_1=\mathcal E$ and $\|\tilde a-\hat a\|_1=|\hat Z-Z|/Z=r_Z$. The triangle and reverse triangle inequalities give
\[
  \mathcal E-r_Z\le\|\hat a-a\|_1\le\mathcal E+r_Z\le2\mathcal E,
\]
where $r_Z\le\mathcal E$. Since $\sum_i(\hat a_i-a_i)=0$, for any reference vector $c$ and $R(c)=\max_i\|v_i-c\|_2$,
\[
  \hat o-o=\sum_i(\hat a_i-a_i)(v_i-c),\qquad
  \|\hat o-o\|_2\le R(c)\|\hat a-a\|_1\le2R(c)\mathcal E.
\]
More generally, for any nonnegative implied weights with $\hat Z>0$,
\[
\hat Z(\hat o-o)=\sum_i(\hat w_i-w_i)(v_i-o),
\]
since $\sum_iw_i(v_i-o)=0$. Therefore
\[
  \|\hat o-o\|_2\le\hat Z^{-1}\sum_i|\hat w_i-w_i|\,\|v_i-o\|_2
  \le\frac Z{\hat Z}\mathcal E R(o).
\]
Since $|\hat Z-Z|\le Z\mathcal E$, for $\mathcal E<1$ we have $Z/\hat Z\le(1-\mathcal E)^{-1}$.

The identity in Lemma~\ref{lem:identity} also has a covariance/Jensen form. Writing $S_b=\sum_{i\in b}e^{z_i}$ and defining covariance with the uniform measure on cluster members,
\[
  \hat Z(\hat o-o)=-\sum_{b\in\mathcal P}\Bigl[
    n_b\,\mathrm{Cov}_{i\in b}(e^{z_i},v_i)
  +(S_b-\hat M_b)(\bar v_b-o)\Bigr].
\]
For a single tail $T$, let
$\tilde v_T=S_T^{-1}\sum_{i\in T}w_iv_i$ and $\hat M_T=Ne^{\bar z_T}$.
The same identity becomes
\[
\hat Z_{\mathrm p}(\hat o_{\mathrm p}-o)
=
-S_T(\tilde v_T-\bar v_T)
-(S_T-\hat M_T)(\bar v_T-o),
\]
whereas discarding gives
\[
Z_X(\hat o_{\mathrm d}-o)
=-S_T(\tilde v_T-o).
\]
Thus similar tokenwise weight discrepancy need not imply similar output error: proxying retains the tail mean-value contribution, leaving only its value-direction and mass mismatches.

For each residual cluster define $G_b=\log(n_b^{-1}\sum_{i\in b}e^{z_i-\bar z_b})$. Mean-logit proxying gives $S_b=\hat M_be^{G_b}\ge\hat M_b$ by Jensen's inequality, hence the exact signed normalizer deficit is
\[
  Z-\hat Z=\sum_{b\in\mathcal P}(S_b-\hat M_b)
  =\sum_{b\in\mathcal P}\hat M_b(e^{G_b}-1)\ge0.
\]
This deficit is bounded by the exact absolute discrepancy $Z\mathcal E=\sum_b\varepsilon_b$ of Eq.~(\ref{eq:identity}) and interprets the signed Jensen alternative below; the deployed \mgate{} estimates the absolute discrepancy. The covariance/Jensen decomposition also supports the value-aware alternative in App.~\ref{app:errgate}. The value-free discrepancy needs no small-covariance assumption: Eq.~(\ref{eq:lemma-identity}) and the triangle inequality give Proposition~\ref{prop:bound} directly. Exact-read tokens contribute nothing, and reading a whole cluster removes its nonnegative discrepancy term; partial reads require recomputing the residual statistics.

\begin{proposition}[deterministic upper bound]\label{prop:bound}
  With $D_b=n_b^{-1}\sum_{i\in b}|e^{\delta_i}-1|$,
  \begin{equation}
    \begin{aligned}
      \|\hat o-o\|_2
      &\le\sum_{b\in\mathcal P}\hat a_b^{\mathrm{virt}}V_bD_b\\
      &\le\sum_{b\in\mathcal P}\hat a_b^{\mathrm{virt}}V_b\bigl(e^{\Delta_b^+}-e^{-\Delta_b^-}\bigr)
      \le\sum_{b\in\mathcal P}\hat a_b^{\mathrm{virt}}V_be^{\Delta_b^+}.
    \end{aligned}
    \label{eq:bound}
  \end{equation}
\end{proposition}
\begin{proof}
  Apply the triangle inequality to Eq.~(\ref{eq:lemma-identity}) and bound $\|v_i-o\|_2\le V_b$:
  \[
    \|\hat o-o\|_2\le\hat Z^{-1}\sum_bV_be^{\bar z_b}\sum_{i\in b}|e^{\delta_i}-1|
    =\sum_b\hat a_b^{\mathrm{virt}}V_bD_b.
  \]
  On $[-\Delta_b^-,\Delta_b^+]$, $|e^\delta-1|\le\max(e^{\Delta_b^+}-1,1-e^{-\Delta_b^-})\le e^{\Delta_b^+}-e^{-\Delta_b^-}$, giving the remaining inequalities.
\end{proof}
The bound factorises per cluster into proxy mass ($\hat a_b^{\mathrm{virt}}\propto n_be^{\bar z_b}$, set by the mean logit), value radius $V_b$, and dispersion of the true weights around the proxy weight ($D_b$, set by the logit spread). Both the mean and the spread enter as factors of the same term.

\begin{corollary}[small dispersion under a moment condition]\label{cor:small}
  If the within-cluster deviations satisfy $\frac1{n_b}\sum_{i\in b}e^{\lambda\delta_i}\le e^{\lambda^2\sigma_b^2/2}$ for $\lambda\in\{1,2\}$, then $D_b\le\sqrt{e^{2\sigma_b^2}-1}$ and
  \begin{equation}
    \begin{aligned}
      \|\hat o-o\|_2&\le\sum_{b\in\mathcal P}\hat a_b^{\mathrm{virt}}V_b\sqrt{e^{2\sigma_b^2}-1}
      \approx\sqrt2\sum_{b\in\mathcal P}\hat a_b^{\mathrm{virt}}V_b\sigma_b
      && (\sigma_b\ll1),\\
      0&\le\frac{S_b}{\hat M_b}-1\le e^{\sigma_b^2/2}-1.
    \end{aligned}
    \label{eq:bound-mgf}
  \end{equation}
\end{corollary}
\begin{proof}
  Cauchy--Schwarz gives $D_b\le(\frac1{n_b}\sum_i(e^{\delta_i}-1)^2)^{1/2} =(\overline{e^{2\delta}}-2\overline{e^\delta}+1)^{1/2}$. Jensen gives $\overline{e^\delta}\ge e^{\bar\delta}=1$, and the moment condition gives $\overline{e^{2\delta}}\le e^{2\sigma_b^2}$. The mass-deficit statement uses the same condition at $\lambda=1$.
\end{proof}
This moment condition is an additional tail-shape assumption; it is not implied by the empirical variance alone and can fail for clusters with rare positive-logit outliers.
Under the stated moment condition, the output bound is first order in small $\sigma_b$, whereas the bound on missing cluster mass is second order. Under the full query metric, the expected dispersion is $\mathbb E_q[\sigma_b^2]=s^2\,\mathrm{tr}(\mathrm{Cov}_b(K)\mathbb E[qq^\top])$. Minimising $\sum_bn_b\mathbb E_q[\sigma_b^2]$ is the $k$-means objective
under $M=\mathbb E[qq^\top]$ (\S\ref{sec:partition}); it therefore
minimises a token-count-weighted dispersion surrogate, not the mass- and value-weighted output bound itself.

\begin{corollary}[Gaussian population estimate of dispersion]\label{cor:jgap}
  Model centered within-cluster deviations by $\delta\sim\mathcal N(0,\sigma_b^2)$. For this population,
  \begin{equation}
    \begin{aligned}
      \mathbb E[e^\delta]-1&=e^{\sigma_b^2/2}-1,\\
      D_{\mathrm G}(\sigma_b):=\mathbb E|e^\delta-1|
      &=e^{\sigma_b^2/2}\bigl(2\Phi(\sigma_b)-1\bigr)
      \approx\sqrt{2/\pi}\,\sigma_b\quad(\sigma_b\ll1),
    \end{aligned}
    \label{eq:closedforms}
  \end{equation}
  where $\Phi$ is the standard normal c.d.f. Finite-cluster $D_b$ and $S_b/\hat M_b-1$ need not equal these population expressions. The exact, Gaussian-model and deployed diagonal quantities are respectively
  \[
    \varepsilon_b=\hat M_bD_b,\qquad
    \varepsilon_b^{\mathrm G}=\hat M_bD_{\mathrm G}(\sigma_b),\qquad
    \hat\varepsilon_b=\hat M_bD_{\mathrm G}(\hat\sigma_b),
  \]
  where $\sigma_b^2=s^2q^\top\mathrm{Cov}_b(k)q$ and $\hat\sigma_b^2=s^2\sum_dq_d^2\mathrm{Var}_b(k_d)$. The last expression is Eq.~(\ref{eq:ucb}). Thus $Z\mathcal E=\sum_b\varepsilon_b$ exactly, while $\sum_b\hat\varepsilon_b$ estimates that quantity before allocated reads. No assumption about independence of weights and values is used.
\end{corollary}
\begin{proof}
  For $\delta\sim\mathcal N(0,\sigma^2)$, the moment generating function gives $\mathbb Ee^\delta=e^{\sigma^2/2}$. Also $\mathbb E|e^\delta-1|=2\mathbb E(e^\delta-1)^+-\mathbb E(e^\delta-1)$ and $\mathbb E[e^\delta\mathbf1\{\delta>0\}]=e^{\sigma^2/2}\Phi(\sigma)$, giving Eq.~(\ref{eq:closedforms}); its small-$\sigma$ form is $2\varphi(0)\sigma$. The identities for $\varepsilon_b$ and $Z\mathcal E$ follow from Eq.~(\ref{eq:identity}); the diagonal substitution defines the deployed score estimate.
\end{proof}

\paragraph{Signed Jensen-gap alternative.}
For a residual proxy cluster the exact normalizer deficit is $\Delta_b=S_b-\hat M_b=\hat M_b(e^{G_b}-1)$, and its Gaussian-diagonal estimate is $\hat\Delta_b=\hat M_b(e^{\hat\sigma_b^2/2}-1)$. This signed mass-deficit quantity is useful for interpretation and ranking ablations, but is distinct from the absolute discrepancy estimate deployed in Eq.~(\ref{eq:ucb}). The latter is first order in small $\sigma_b$, while the modeled deficit is second order.

\paragraph{From cluster discrepancy to a token budget.}
For a fixed partition and a budget of $m$ complete clusters, reading the $m$ clusters with largest exact $\varepsilon_b$ removes the largest nonnegative terms from $Z\mathcal E$. For partial reads, \mgate{} uses the estimated score and the proportional quota of Eq.~(\ref{eq:quota}),
\[
  q_b=\min\!\left(n_b,\lambda\frac{\hat\varepsilon_b}{\max_{b'}\hat\varepsilon_{b'}}\right),
  \qquad \sum_bq_b=E,
\]
with $\lambda$ set by water filling and integer allocation. The \qgs{} chooses the $q_b$ highest-scoring tokens in each cluster, and proxy statistics are recomputed over unread members (App.~\ref{app:errgate}). This partial-token allocation is a deployment design choice motivated by the measured concentration of discrepancy; the complete-cluster argument does not prove it optimal.

\begin{corollary}[alternative maximum-based ranking heuristic]\label{cor:rank}
  Under the sub-Gaussian maximum heuristic $\Delta_b^+\lesssim\sigma_b\sqrt{2\log n_b}$, the last form of Eq.~(\ref{eq:bound}) becomes
  \begin{equation}
    \|\hat o-o\|_2\lesssim\frac1{\hat Z}\sum_{b\in\mathcal P}V_bn_b
    \exp\!\bigl(\bar z_b+c_b\sigma_b\bigr),\qquad c_b=\sqrt{2\log n_b}.
    \label{eq:bound-max}
  \end{equation}
  Large numerator contributions motivate a ranking heuristic, but removing their largest terms does not in general minimise this normalised bound: $\hat Z$ also changes with the exact set. Dropping the unavailable $V_b$, treating $n_b$ as comparable across clusters and fixing $c$ gives the alternative score $\bar z_b+c\sigma_b$, distinct from Eq.~(\ref{eq:ucb}).
\end{corollary}
Both proxy mass $n_be^{\bar z_b}$ and within-cluster dispersion matter: a cluster with $\sigma_b=0$ is represented exactly by its proxy, while a cluster with tiny proxy mass contributes little at a fixed dispersion. The constant $c$ in the maximum heuristic is not a calibrated tail bound, and heavy-tailed clusters can degrade its coverage. The maximum-based heuristic and the deployed absolute score use different functions of the same cached moments.

\subsection{\mgate{}: the estimate and the hierarchical budget}
\label{app:errgate}
\paragraph{The estimate.}
Corollary~\ref{cor:jgap} distinguishes the finite-cluster discrepancy $\varepsilon_b=\hat M_bD_b$ from the Gaussian model $\varepsilon_b^{\mathrm G}=\hat M_bD_{\mathrm G}(\sigma_b)$ and the diagonal estimate $\hat\varepsilon_b=\hat M_bD_{\mathrm G}(\hat\sigma_b)$ used in Eq.~(\ref{eq:ucb}). Since $2\Phi(\hat\sigma_b)-1=\mathrm{erf}(\hat\sigma_b/\sqrt2)$, the deployed score is
\[
  \hat\varepsilon_b=\hat M_b e^{\hat\sigma_b^2/2}\,
  \mathrm{erf}(\hat\sigma_b/\sqrt2),\qquad
  \hat\sigma_b^2=s^2\sum_dq_d^2\mathrm{Var}_b(k_d).
\]
Its units are unnormalised mass, and the estimates add to an estimate of $Z\mathcal E$, not to its exact value. The rule uses a few elementwise operations and one error function on a $(B,H,1,C)$ tensor. The factor $e^{\bar z_b}$ retains the sign-sensitive ordering of the mean logit; squaring the mean would erase this distinction.

\paragraph{Alternative estimates.}
The signed form $\hat\Delta_b=\hat M_b(e^{\hat\sigma_b^2/2}-1)$ estimates the Jensen mass deficit, in which positive and negative weight deviations cancel. It is second order in small dispersion, whereas $\hat\varepsilon_b$ is first order. With the same split and residual statistics, the signed alternative scores 86.28 at a 1\% read against \OURSONE{} for Eq.~(\ref{eq:ucb}). A value-aware alternative estimates output error instead. Define the exact weighted cluster mean $\tilde v_b=\sum_{i\in b}e^{z_i}v_i/S_b$ and the residual
\[
  r_b=\sum_{i\in b}(e^{z_i}-e^{\bar z_b})(v_i-o)
  =(S_b-\hat M_b)(\tilde v_b-o)+\hat M_b(\tilde v_b-\bar v_b).
\]
Then Lemma~\ref{lem:identity} gives $\hat Z(\hat o-o)=-\sum_b r_b$. Cauchy--Schwarz gives $\|\tilde v_b-\bar v_b\|_2\le\mathrm{CV}_b(e^z)\nu_b$, where $\nu_b^2=\mathrm{tr}\,\mathrm{Cov}_b(v)$. Under the Gaussian model the population coefficient of variation is $\sqrt{e^{\sigma_b^2}-1}$. Substituting the diagonal variance and a partial exact-token output $\hat o_X$ yields the alternative heuristic
\[
  \hat\varepsilon_b^{\,\mathrm{out}}=n_be^{\bar z_b}
  \Bigl[(e^{\hat\sigma_b^2/2}-1)\|\bar v_b-\hat o_X\|_2
  +\sqrt{e^{\hat\sigma_b^2}-1}\,\nu_b\Bigr].
\]
This score also replaces the mass-channel value offset by $\bar v_b-\hat o_X$; it is not the exact residual or a certified bound. It needs the cached value spread and the partial output, and scores 86.04 with the same split. The three variants are within one standard error; the deployed score keeps its estimate value-free and uses only the mean and diagonal variance tokens. This comparison is not a small-covariance assumption in Proposition~\ref{prop:bound}.

\paragraph{Why retain the uncorrected proxy.}
For a constant weight $u$, $\sum_{i\in b}|w_i-u|$ is minimized by a median of the weights. Under $z=\bar z_b+\delta$ with $\delta\sim\mathcal N(0,\sigma_b^2)$, the lognormal population median is $e^{\bar z_b}$, the mean-logit proxy weight. We use this uncorrected weight as a model-based design choice, not a finite-cluster or output-error guarantee.

\paragraph{Diagonal moments and discrepancy concentration.}
The mean and variance tokens cost 1.5 KV rows per cluster per query, both charged to the read. In the $C{=}64$ QGC diagnostic of App.~\ref{app:qgsprotocol}, diagonal and full-variance model estimates agree at Spearman \ESTRHODF{} and recover the eight clusters with the largest measured exact discrepancy at similar rates (\GATEOVDT{}, \GATEOVFT{}). These measurements assess the diagonal substitution, not a general variance inequality.

In the discrepancy-concentration diagnostic, clusters with at least 32 of the exact top-1\% tokens are \NTOPBIGFRAC{} of clusters and carry \NTOPBIGSHARE{} of the exact discrepancy; the \NTOPZEROFRAC{} of clusters without such tokens carry \NTOPZEROSHARE{}. This concentration motivates, but does not prove optimal, the proportional quota of \S\ref{sec:rank}.

\paragraph{Statistics of the remaining members.}
Before quota allocation, the initially selected sink, local and QGS top-$k$ tokens are subtracted from the cached pooled count, first moment and second moment. Thus $n_b$, $\bar z_b$ and $\hat\sigma_b^2$ used by ErrorGate are statistics of the candidate residual members, not of the original cluster.
After cluster $b$ reads $q_b$ additional tokens, these tokens are subtracted again to form the final proxy statistics, so no token is counted twice. Lemma~\ref{lem:identity} and Proposition~\ref{prop:bound} apply to this final residual state, whereas Eq.~(\ref{eq:ucb}) scores the pre-allocation residual.

\paragraph{Hierarchical budget.}
Given the estimates, water filling uses $q_b=\min(n_b,\lambda w_b)$ with $w_b=\hat\varepsilon_b/\max_{b'}\hat\varepsilon_{b'}$ and $\lambda$ chosen to meet the token budget $E$. After flooring, the remainder goes one token at a time to the highest-ranked clusters with capacity, so $\sum_bq_b=E$ when the budget is feasible. Within a cluster, the $q_b$ tokens are its highest under the \qgs{} score of Eq.~(\ref{eq:qgs}). A uniform cap $t$, used in the earlier comparison, spends the same $E$ on approximately $E/t$ clusters in equal parts; the proportional split instead gives most reads to the few clusters with the largest estimates. At 1\%, it reaches about 27 clusters per row with the largest quota near 60 tokens. Its total budget is fixed per row.  Neither complete-cluster optimality nor the concentration diagnostic proves the proportional partial-token allocation optimal.

\subsection{Closed forms of the sampling and proxy errors}
\label{app:proxyformula}
This subsection derives Eq.~(\ref{eq:tail-errors}), gives the exact finite-tail proxy discrepancy, and records the protocol behind Figure~\ref{fig:formula}.

\paragraph{Setup.}
Fix a query and a nonempty exact set $X$, with nonempty tail $T=[n]\setminus X$, $N=|T|$, $Z_X=\sum_{i\in X}w_i$, $S_T=\sum_{i\in T}w_i$ and $\tau=S_T/Z$, where $Z=Z_X+S_T$. All expectations below condition on this fixed row and tail. Exact-read tokens contribute nothing to $Z\mathcal E$.

\paragraph{Discarding and random sampling.}
Discarding sets $\hat w_j=0$ on $T$, hence $\mathcal E_{\mathrm d}=S_T/Z=\tau$. For a uniformly chosen subset $B\subseteq T$ of size $m$ without replacement, set $\eta=N/m$. The sampling weights are $\hat w_j=\eta w_j$ on $B$ and zero on $T\setminus B$, giving
\[
Z\mathcal E_{\mathrm s}
=\sum_{j\in B}(\eta-1)w_j+\sum_{j\in T\setminus B}w_j.
\]
Each indicator $I_j=\mathbf1_B(j)$ has expectation $m/N$, so
\[
\mathbb E[Z\mathcal E_{\mathrm s}]
=\bigl[(\eta-1)m/N+(1-m/N)\bigr]S_T
=2(1-m/N)S_T.
\]
Dividing by the fixed $Z$ gives Eq.~(\ref{eq:tail-errors}). The spread of tail weights affects the variance of this discrepancy, not its expectation at fixed $S_T$, $Z$ and $m/N$.

\paragraph{Normalizer uncertainty and missed tokens.}
The sampling normalizer $\hat Z_{\mathrm s}=Z_X+\eta\sum_{j\in B}w_j$ is unbiased for $Z$. For $N>1$, let $\bar w=S_T/N$, $\sigma_w^2=N^{-1}\sum_{j\in T}(w_j-\bar w)^2$ and $\mathrm{CV}_T=\sigma_w/\bar w$. Sampling without replacement gives
\[
\frac{\sqrt{\mathbb E[(\hat Z_{\mathrm s}-Z)^2]}}{Z}
=\tau\,\frac{\mathrm{CV}_T}{\sqrt m}
 \sqrt{\frac{N-m}{N-1}}.
\]
Equivalently, with $N_{\mathrm{eff}}=S_T^2/\sum_{j\in T}w_j^2$,
$\mathrm{CV}_T^2=N/N_{\mathrm{eff}}-1$.
Thus the uncertainty grows both as the sample shrinks and as the residual mass concentrates on fewer effective tokens.
This measures uncertainty in total mass, not its tokenwise allocation. At fixed tail statistics it grows as the sample shrinks. A given unread token is included with probability $m/N$; missing a large-weight token can affect both the sampled mass and value direction. At a 1\% read behind HashAttention, the reported inclusion probability of the heaviest unread token is \PIPEHITHATONE{} in the \PIPENROWS{}-row diagnostic (App.~\ref{app:pipediag}). Matched-read experiments also change the tail, so they are not fixed-tail sweeps in $m$ alone.

\paragraph{One tail proxy.}
Let $g_T=e^{\bar z_T}$ and $\bar v_T=N^{-1}\sum_{i\in T}v_i$. Equation~(\ref{eq:tail-weights}) assigns $g_T/\hat Z_{\mathrm p}$ to each tail token, equivalently one virtual token of weight $Ng_T/\hat Z_{\mathrm p}$ and value $\bar v_T$:
\[
\sum_{i\in T}\hat a_i^{\mathrm p}v_i
=\frac{Ng_T}{Z_X+Ng_T}\bar v_T.
\]
Define the finite-tail quantities
\[
G_T=\log\!\left(\frac1N\sum_{j\in T}e^{z_j-\bar z_T}\right),\qquad
D_T=\frac1N\sum_{j\in T}|e^{z_j-\bar z_T}-1|.
\]
Since $Z\mathcal E_{\mathrm p}=Ng_TD_T$ and $Ng_T=S_Te^{-G_T}$, the exact proxy discrepancy is
\begin{equation}
\mathcal E_{\mathrm p}=\tau e^{-G_T}D_T.
\label{eq:proxyerr}
\end{equation}
Jensen's inequality gives $G_T\ge0$, and the relative normalizer deficit is exactly $(Z-\hat Z_{\mathrm p})/Z=\tau(1-e^{-G_T})$. The proxy has no sampling variance. Under the Gaussian population model, $G_{\mathrm G}=\sigma_z^2/2$ and $D_{\mathrm G}(\sigma_z)=e^{\sigma_z^2/2}(2\Phi(\sigma_z)-1)$ (Corollary~\ref{cor:jgap}), hence
\[
\mathcal E_{\mathrm p}^{\mathrm G}
=\tau e^{-\sigma_z^2/2}D_{\mathrm G}(\sigma_z)
=\tau\,\mathrm{erf}(\sigma_z/\sqrt2).
\]
This yields the Gaussian prediction in Eq.~(\ref{eq:tail-errors}), not a guarantee for every finite tail. Empirical $G_T$ and $D_T$ need not equal the population expressions.

\paragraph{Protocol.}
The rows are those of App.~\ref{app:qgsprotocol}, \CDNROWS{} decode rows
of Llama-3.1-8B-Instruct over the RULER-32K HARD tasks, three requests
each, every layer and head and the first eight generated tokens.
For each total read $\rho\in\{0.1,0.5,1,3,5,10\}\%$, each decode row
uses its valid cache length $n$ and reserves $n_{\mathrm{base}}=6$
sink and local tokens. The remaining budget
$\mathrm{free}=\rho n-n_{\mathrm{base}}$ is split 50/50: the sampling
treatment reads the selector's top half exactly and draws the other half
uniformly from the residual tail.

The $m$ values reported in Table~\ref{tab:proxyformula} are the mean
realized sample-mask counts over decode rows. Each treatment therefore
has its own residual tail and tail mass $\tau$. Per selector and read,
$\tau_{\mathrm{s}}$ and $\tau_{\mathrm{p}}$ are row-averaged tail-mass statistics, while the reported $m/N$ is an aggregate sample-to-tail ratio rather than the mean of the rowwise ratios.
The quantity $\sigma_z=\sqrt{\langle\sigma_{z,i}^2\rangle}$ is computed over the proxy tails. The reported predictions are plug-in values evaluated on
these aggregate statistics. The measured columns instead average
discrepancies computed directly from the raw decode rows, with one draw
per row for sampling. The discard measurement follows a tail-selection
path that is not rowwise identical to that underlying its plug-in
prediction. Thus the prediction/measurement pairs are descriptive
comparisons rather than rowwise-matched validation.

\begin{table}[h]
  \centering
  \caption{ Inputs and outputs of Fig.~\ref{fig:formula}. $\mathcal E$ is the exact-normalizer-scaled unnormalised weight discrepancy. Predictions (f) are plug-in values from aggregate tail statistics, whereas measurements (m) are averages computed from raw decode rows; they are therefore not rowwise-paired. The proxy prediction additionally uses the Gaussian population model.}
  \label{tab:proxyformula}
  
  \setlength{\tabcolsep}{4pt}
  \begin{tabular}{lrrrrrrrr}
    \toprule
    selector & read & $\tau_{\mathrm{s}}$ & $\tau_{\mathrm{p}}$ & $\sigma_z$ & $m$ & $\mathcal E_{\mathrm{s}}$ (f / m) & $\mathcal E_{\mathrm{p}}$ (f / m) & $\mathcal E_{\mathrm{d}}$ (f / m) \\
    \midrule
    HashAttention & 10\% & 0.212 & 0.176 & 2.36 & 1585 & 0.401 / 0.398 & 0.173 / 0.171 & 0.176 / 0.176 \\
    & 5\% & 0.245 & 0.212 & 2.41 & 791 & 0.477 / 0.480 & 0.208 / 0.206 & 0.212 / 0.212 \\
    & 3\% & 0.268 & 0.236 & 2.43 & 473 & 0.527 / 0.530 & 0.233 / 0.230 & 0.236 / 0.236 \\
    & 1\% & 0.315 & 0.286 & 2.47 & 156 & 0.628 / 0.623 & 0.282 / 0.280 & 0.286 / 0.286 \\
    & 0.5\% & 0.345 & 0.316 & 2.48 & 76 & 0.687 / 0.684 & 0.312 / 0.310 & 0.316 / 0.316 \\
    & 0.1\% & 0.406 & 0.386 & 2.50 & 13 & 0.812 / 0.855 & 0.381 / 0.380 & 0.386 / 0.385 \\
    \midrule
    oracle & 10\% & 0.049 & 0.032 & 2.10 & 1585 & 0.093 / 0.093 & 0.030 / 0.027 & 0.032 / 0.032 \\
    & 5\% & 0.067 & 0.049 & 2.24 & 791 & 0.131 / 0.131 & 0.048 / 0.043 & 0.049 / 0.049 \\
    & 3\% & 0.081 & 0.062 & 2.32 & 473 & 0.159 / 0.159 & 0.061 / 0.056 & 0.062 / 0.062 \\
    & 1\% & 0.109 & 0.092 & 2.42 & 156 & 0.218 / 0.218 & 0.090 / 0.085 & 0.092 / 0.091 \\
    & 0.5\% & 0.128 & 0.110 & 2.45 & 76 & 0.256 / 0.256 & 0.108 / 0.104 & 0.110 / 0.110 \\
    & 0.1\% & 0.195 & 0.166 & 2.49 & 13 & 0.390 / 0.390 & 0.164 / 0.160 & 0.166 / 0.165 \\
    \bottomrule
  \end{tabular}
\end{table}

\paragraph{What the weight error does and does not say.}
The discrepancy $\mathcal E$ counts absolute differences of implied
unnormalised weights and is not the normalized attention distance
$\|\hat a-a\|_1$; Appendix~\ref{app:bound} gives their relation.
The prediction/measurement pairs in Table~\ref{tab:proxyformula} are
descriptive rather than a calibration test: the predictions use
aggregate tail statistics, whereas the measurements are computed
rowwise, and the discard columns additionally use slightly different
tail-selection conditions. In particular, agreement of the proxy
column does not validate the Gaussian tail model. The metric also
ignores values, so its ordering need not match output error;
Figure~\ref{fig:tailcurves} reports the corresponding output comparison
on a separate row set. In this subsection $m$ is the sample size;
elsewhere in the appendix $m$ counts expanded clusters.

\section{Attention quality and diagnostics}

\subsection{Quality and error across read densities}
\label{app:qualityerror}
\begin{figure}[htbp]
  \centering
  \includegraphics[width=\linewidth]{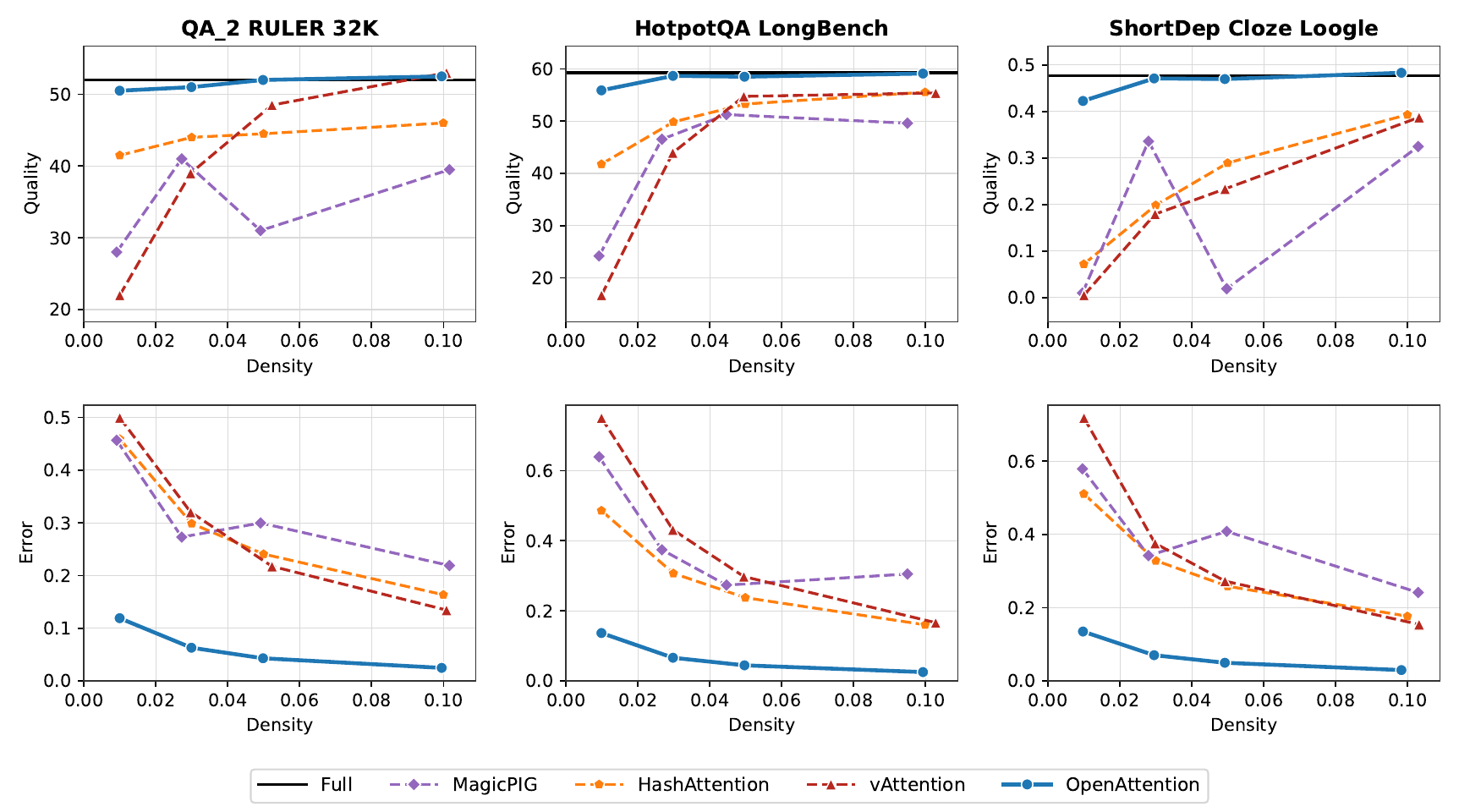}
  \caption{Quality and error versus read density on representative RULER-32K, LongBench, and LooGLE tasks.}
  \label{fig:qualityerror}
\end{figure}
Figure~\ref{fig:qualityerror} plots task quality and error as the read density increases from 1\% to 10\% for RULER-32K QA\_2, LongBench~\citep{bai2024longbench} HotpotQA~\citep{yang2018hotpotqa}, and LooGLE~\citep{li2024loogle} ShortDep Cloze. Across these three tasks, \ours{} stays close to dense attention quality while showing the lowest error among the sparse methods at the plotted densities. The separation is clearest at low density, where the baselines' errors rise and their task quality falls. These curves complement the fixed-budget results in the main text by showing how both measures change with the read budget.

\subsection{Protocol of the decode-row diagnostic}
\label{app:pipediag}

\begin{figure}[h]
  \centering
  \includegraphics[width=\linewidth]{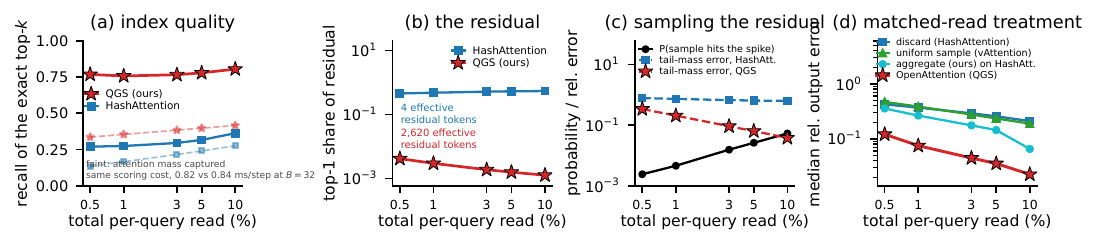}
  \caption{ One pass over \PIPENROWS{} RULER-32K HARD decode rows (Llama-3.1-8B, all layers and heads of 21 requests) against the \emph{total} read. (a) Recall of the exact top-$k$ by the HashAttention signature and by the deployed \qgs{} (64 vs.\ 32 bits/token/head at the same per-step cost, $0.84$ vs.\ $0.82$\,ms at $B{=}32$), with the captured mass in faint lines. (b) Median share of the unread mass held by its largest token, annotated with the residual's participation ratio. (c) Probability that \vatt{}'s sample contains that token, and the relative error of the tail mass it estimates. (d) Median relative output error of four treatments of the same residual at the \emph{same} read, which discard it, sample it, aggregate it into clusters, or apply the deployed \ours{}. The protocol is given in App.~\ref{app:pipediag}.}
  \label{fig:pipeline}
\end{figure}
Figure~\ref{fig:pipeline} is one pass of \texttt{run\_v37\_tail\_diag.py} over the RULER-32K HARD tasks, three requests each, recording every layer, every head and the first eight generated tokens, which gives \PIPENROWS{} rows. Dense output is returned at every call, so the generation trajectory is the model's own. Within a row the two indices, the five reads and the four treatments all see the same cache and the same query, and each treatment is charged the same \emph{total} read $\rho n$. Sink and local hold the deployed few-token floor (0.01\% each), and the remainder is spent in one of four ways. Treatment (i) spends it on the selector top-$k$ only. Treatment (ii) spends half on the top-$k$ and half on a random sample of $m$ residual tokens, rescaled by $N/m$ and merged into the same softmax, which is \vatt{}'s fixed 50/50 split (App.~\ref{app:vattprotocol}). Treatment (iii) reads $C$ clusters at $1.5$ rows each and the rest on the top-$k$, with every cluster a proxy. Treatment (iv) uses $C$ clusters, a $\sim$9-token top-$k$ and $E$ tokens expanded by Eq.~(\ref{eq:ucb}) with the uniform per-cluster cap $t$ of earlier versions; the deployed quotas of App.~\ref{app:errgate} change the 1\% mean by less than one standard error. The clusters are rebuilt inside the diagnostic as at deployment (keys aligned as $\phi_i=\Lambda_r^{1/2}U_r^\top k_i$ with $r{=}16$ under the shrunk $M$, four Lloyd iterations in bf16, once per request, new keys joining the nearest centroid) and the codes and cluster assignments of tokens generated during decoding are extended incrementally, as at deployment. The \qgs{} is the deployed one ($r{=}32$, int8, one $U$ per KV head).

\subsection{Protocol of the selector and aggregation diagnostics}
\label{app:qgsprotocol}
Fig.~\ref{fig:errgate} comes from one pass of \texttt{run\_v49\_cluster\_diag.py} over the RULER-32K HARD tasks, three requests each, every layer and head and the first eight generated tokens (\CDNROWS{} rows), with dense output returned so the trajectory is the model's own, as in App.~\ref{app:pipediag}. \emph{Partitions.} The tail is everything but sink, local and the \qgs{} top-1\%, and every cluster is aggregated, so the error isolates the partition. Key-only $k$-means runs the same four Lloyd iterations as ours on the centred raw keys, one partition per KV head, and adjacency cuts the cache into $C$ equal runs. The reported dispersion is the within-cluster standard deviation of the exact logits, averaged over clusters with weights equal to their true attention mass. \emph{Clusters.} Fig.~\ref{fig:errgate} and the error shares by top-$k$ content reported in App.~\ref{app:errgate} use the deployed 1\% partition ($C{=}64$) with only the $\sim$9-token \qgs{} top-$k$ exact, on every fourth head (2.2M clusters). The exact variance is the within-cluster variance of the true logits, and the true error of a cluster is its weight error $\hat M_bD_b/Z=\sum_{i\in b}|e^{z_i}-e^{\bar z_b}|/Z$ (Eq.~(\ref{eq:identity})); the partition error is the same sum over all clusters with every cluster a proxy; on these rows the attention weighted within cluster logit std at $C{=}64$ is \PARTDISPADJ{} for positional clusters, \PARTDISPKEY{} for $k$-means on the keys alone and \PARTDISPOURS{} for \qgc{}, and the weight error follows (\PARTERRADJ{}, \PARTERRKEY{}, \PARTERROURS{}), so \qgc{} at $C{=}64$ beats key only clusters at $C{=}1024$. Over all rows, the eight clusters with the largest $\hat\varepsilon_b$ hold \TOPEIGHTERR{} of the aggregation error and \TOPEIGHTTOPK{} of the exact top-1\% tokens left in the tail, and $\hat\varepsilon_b$ orders the clusters by their true error at Spearman \ESTRHOTRUE{}.

\section{\oursflash{}: \ours{} on FlashAttention-2}
\label{app:kernel}
Algorithm~\ref{alg:flash} is the decode attention step of \S\ref{sec:kernel}. A FlashAttention-2 decode kernel keeps, per program, a running maximum, a running normaliser and an output accumulator, and folds one tile of keys and values at a time into them; \oursflash{} keeps that loop, written in Triton, and changes only what the tiles hold. Phases 1 and 2 gather the exact tokens, the \qgs{} top-$k$ and the tokens \mgate{}'s quotas read from the expanded clusters, from scattered rows of the cache straight into SRAM, where a gather in PyTorch writes the gathered tile to DRAM and reads it back. Phase 3 streams the clusters as virtual tokens, the mean key in place of a key, the mean value in place of a value and the occupancy of the unread members as $+\log n_b'$ on the logit, so the tail costs one more tile loop rather than a second attention pass. The statistics of the unread members are prepared before the launch by subtracting the tokens the query reads from the cached pooled sums (App.~\ref{app:errgate}), and a cluster read whole carries $\log 0=-\infty$, so the kernel needs no membership test. The index lists are padded to a fixed length with masked slots, so the work per step is fixed, with no data dependent loop bound and no host synchronisation.

\begin{algorithm}[h]
  \caption{ \oursflash{}: one decode step of the residual-proxy estimator as a FlashAttention-2 online softmax pass. One program per (sequence, query head); the $G{=}4$ query heads of a KV head read the same cache; $T{=}64$ is the tile size and $s=d^{-1/2}$.}
  \label{alg:flash}
  \begin{algorithmic}[1]
    \Require query $q$; cache $K,V$ of the KV head; the index list $\mathcal S$ of the selected tokens (the \qgs{} top-$k$) and $\mathcal X$ of the tokens the expanded clusters read (the \mgate{} quotas $q_b$ of \S\ref{sec:rank}); per cluster the statistics of its \emph{unread} members, $(\bar k_b',\bar v_b',n_b')$, obtained from the cached pooled sums by subtracting the tokens in $\mathcal S\cup\mathcal X$, with $n_b'{=}0$ for a cluster read whole
    \Ensure $\hat o$ of the residual-proxy estimator
    \State $m\gets-\infty$,\ $L\gets0$,\ $o\gets\mathbf 0$ \Comment{running maximum, normaliser and accumulator, fp32}
    \For{each tile of $T$ indices $i$ in $\mathcal S$, then in $\mathcal X$} \Comment{phases 1 and 2: exact tokens}
    \State gather the rows $k_i,v_i$ from the cache into SRAM, padding slots masked
    \State $z_i\gets s\,q^\top k_i$
    \State $(m,L,o)\gets\textsc{Update}(m,L,o,\{z_i\},\{v_i\})$
    \EndFor
    \For{each tile of $T$ clusters $b$} \Comment{phase 3: proxies as virtual tokens}
    \State load $\bar k_b'$, $\bar v_b'$ and $\log n_b'$ \Comment{$\log 0=-\infty$ drops a cluster read whole}
    \State $\ell_b\gets s\,q^\top\bar k_b'+\log n_b'$
    \State $(m,L,o)\gets\textsc{Update}(m,L,o,\{\ell_b\},\{\bar v_b'\})$
    \EndFor
    \State \Return $o/L$
    \Statex
    \Procedure{Update}{$m,L,o,\{z_j\},\{v_j\}$} \Comment{the FlashAttention-2 online softmax step}
    \State $m'\gets\max(m,\max_j z_j)$;\quad $\alpha\gets e^{m-m'}$;\quad $p_j\gets e^{z_j-m'}$
    \State $o\gets\alpha\,o+\sum_j p_jv_j$;\quad $L\gets\alpha L+\sum_j p_j$;\quad $m\gets m'$
    \State \Return $(m,L,o)$
    \EndProcedure
  \end{algorithmic}
\end{algorithm}

\paragraph{Implementation and correctness.}
The kernel runs one program per (sequence, query head), with $G{=}4$ query heads per KV head, tiles of 64 tokens / 64 clusters, 4 warps and fp32 accumulators. Against an fp32 reference the median relative error is $1.6\times10^{-3}$ and the maximum $2.1\times10^{-3}$ over batch-head counts 4--64, including masked expanded clusters. End to end on RULER-32K HARD at 5\%, the kernel scores 85.45 against 85.35 for PyTorch, with per-task differences in $[-0.2,+0.5]$ of mixed sign. Scattered (cluster) expansion versus contiguous expansion costs $+4$--$6\%$ kernel time, so clusters need not be physically re-packed. Each cluster caches $n_b$, $\bar k_b$, $\mathrm{diag}\,\mathrm{Var}_b(k)$ and $\bar v_b$, 385 bf16 scalars, so at the deployed $C{=}64$ the statistics of the whole 32 layer model take 12\,MiB when stored once per KV head, 0.3\% of the 4.0\,GiB KV cache they summarise.

\begin{table}[h]
  \centering
  \caption{ Decode attention step per layer (ms, H100 NVL, 32K, bf16) at 1\% and 3\% reads with no \mgate{} expansion ($E{=}0$), as in Fig.~\ref{fig:density}(c). The \vatt{}/\mpig{} rows are lower bounds charging only their gathers and estimator arithmetic; selection, \mgate{} scoring and residual-statistics updates are excluded for every method. The three models of \S\ref{sec:multimodel} share these attention shapes.  The speedup reported here is therefore of the attention stage and not of the decode step as a whole (see below).}
  \label{tab:latency}
  
  \setlength{\tabcolsep}{4pt}
  \begin{tabular}{lccccc}
    \toprule
    & custom & \multicolumn{2}{c}{1\% read} & \multicolumn{2}{c}{3\% read} \\
    method & kernel & $B{=}1$ & $B{=}32$ & $B{=}1$ & $B{=}32$ \\
    \midrule
    \tablatencyrows
    \bottomrule
  \end{tabular}
\end{table}

\paragraph{What the index scan costs, and why it is excluded.}
The scan that turns a compressed cache into a ranking, and the top-$k$ selection over it, is common to every method in Table~\ref{tab:latency}: \ours{}, HashAttention, \vatt{} and \mpig{} all scan a compressed cache of the same length and take a top-$k$ over it, so excluding it is what makes the sparse rows comparable with one another. It is not what makes them comparable with dense attention. Its cost is method-specific and is not measured in Table~\ref{tab:latency}, so a decode step that charges it gains less than the attention-stage figure. We report the attention stage because that is the part this paper changes, and the $10\times$ refers to this stage only. Bringing the selection itself under the budget is open work.

\section{Budget split and deployed hyper-parameters}
\label{app:fairsplit}

\subsection{Choosing $(C,E,t)$ per budget}
\label{app:denstune}
The grid below was run with the uniform per-cluster cap $t$ and the bound route ranking $\bar z_b+3\sigma_b$ of earlier versions; the deployed \mgate{} replaces the cap by the error proportional quotas of App.~\ref{app:errgate} and keeps the $(C,E)$ the grid selected. Within a budget, $C$, $E$ and the cap $t$ trade against one another at a fixed $\rho$ ($E+1.5C$ is what the accounting leaves once sink, local and the $\sim$9-token top-$k$ are paid), so the triple is chosen per budget on the small grid of Table~\ref{tab:denstune}. The free parameter is really the \emph{share of the read the proxy stream takes}, and the grid moves it while keeping the selector, the partition, the ranking rule and the cap rule fixed. The share that wins grows as the read shrinks, because the sparser the read, the more of it is worth spending on knowing where the mass is rather than on reading tokens.
\begin{table}[h]
  \centering
  \caption{ Per-budget grid over the split between the proxy stream and the expansion (RULER-32K HARD, 200 samples per task; ``proxy'' is $1.5C/\rho n$, the share of the read the cluster statistics take). Everything but $(C,E,t)$ is held fixed. The chosen row is the one Figure~\ref{fig:taskradars} reports.}
  \label{tab:denstune}
  \setlength{\tabcolsep}{3.5pt}
  
  \begin{tabular}{lcccccccccccc}
    \toprule
    $\rho$ & $C$ & $E$ & $t$ & proxy & avg & qa\_1 & qa\_2 & vt & fwe & mk\_2 & mk\_3 & mv \\
    \midrule
    \tabdenstune
    \bottomrule
  \end{tabular}
\end{table}

\section{Results at every budget}
\label{app:densityfull}
Table~\ref{tab:densityfull} gives Figure~\ref{fig:taskradars} per task. Table~\ref{tab:ablationfivefull} gives the per-task results underlying the 5\% component ablation in Table~\ref{tab:ablation}. \ours{} scores \OURSTEN{}, \OURSFIVE{}, \OURSTHREE{} and \OURSONE{} at 10, 5, 3 and 1\%, within 0.25 points of dense down to 3\% and within one point at 1\%, while every baseline degrades steeply; \vatt{} trails plain HashAttention by 2.7, 7.7 and 11.9 points at 5, 3 and 1\% and overtakes it only at 10\%, and \mpig{}, calibrated to the same reads, degrades fastest (App.~\ref{app:magicpig}).

\begin{table}[h]
  \centering
  \caption{ RULER-32K HARD, Llama-3.1-8B-Instruct, 200 samples per task, at a matched \emph{total} per-query read from 10\% to 1\% (``read'' counts the exact rows plus, for \ours{}, the proxy stream of 1.5 rows per cluster, and for \vatt{}, whose sample is adaptive, the measured read); the averages are Figure~\ref{fig:taskradars}. Baselines use the HashAttention selector.}
  \label{tab:densityfull}
  
  \setlength{\tabcolsep}{4pt}
  \resizebox{\linewidth}{!}{%
    \begin{tabular}{llccccccccc}
      \toprule
      budget & method & read & avg & qa\_1 & qa\_2 & vt & fwe & mk\_2 & mk\_3 & mv \\
      \midrule
      \tabdensityours
      \bottomrule
  \end{tabular}}
\end{table}

\begin{table}[h]
  \centering
  \caption{Per-task accuracy for the component ablation of Table~\ref{tab:ablation} at a matched 5\% total read on RULER-32K HARD (200 requests per task).}
  \label{tab:ablationfivefull}
  \setlength{\tabcolsep}{4pt}
  \resizebox{\linewidth}{!}{%
    \begin{tabular}{ccccccccccc}
      \toprule
      proxies & \mgate{} & \qgs{} & avg & qa\_1 & qa\_2 & vt & fwe & mk\_2 & mk\_3 & mv \\
      \midrule
      \tabablationfiverows
      \bottomrule
    \end{tabular}}
\end{table}

\section{Baseline configurations}
\label{app:baselines}
The baselines of \S\ref{sec:results} run inside one framework on the HashAttention selector with the public signature weights. This appendix gives the \mpig{} and \vatt{} configurations that realise each target read under our accounting.

\subsection{MagicPIG configuration}
\label{app:magicpig}
 We use the \mpig{} masker of the \vatt{} authors' framework (random
  signed projections, key centring, the Neyshabur--Srebro maximum inner
  product search (MIPS) transform, a match if any of the $L$ tables
  collides, importance weights $1/p_i$ from the collision probability).
  Here $K$ is the number of hash bits per table and $L$ is the number of
  tables. Its realised density depends on the hash configuration, on the
  query--key geometry and on the context length, so we keep $K$ fixed
  within each sweep and calibrate $L$ to the target read on the benchmark
  itself, accepting a configuration only when its measured read matches the
  budget.

  The \mpig{} points in Figure~\ref{fig:density}(a,b),
  Figure~\ref{fig:taskradars}, and Table~\ref{tab:densityfull} use $K{=}9$,
  with $L{=}8,25,46,100$ at the 1, 3, 5, and 10\% targets, respectively.
  Their measured reads are 1.09, 2.88, 5.13, and 9.97\%. These 32K runs
  evaluate 200 requests per task; the 1\% run used two RTX~4090 GPUs with
  the model split across them, and the other three used one H100 NVL. The
  context-length comparison in Tables~\ref{tab:length}
  and~\ref{tab:lengthfull} is a separate $K{=}10$ sweep at a 3\% target
  read. At 16K, 32K, 64K, and 96K, it uses $L{=}210,300,300,250$ and
  measures reads of 2.99, 3.20, 3.19, and 3.02\%, respectively, with 100
  requests per task.

  Note that \mpig{}'s own evaluation processes the question with
  \emph{dense} attention and applies sampling only to generation, whereas
  the protocol shared by all methods here applies sparse attention to the
  question chunk as well; under this protocol, and without the dense layers
  0 and 16 of the released code, \mpig{} is known to degrade~\citep[App.~C,
  Table~10]{desai2026vattention}.

\subsection{\vatt{} configuration protocol}
\label{app:vattprotocol}
\vatt{}'s accuracy at a given read depends on four parameters, namely the shares of the budget given to the heavy-hitter selector ($f_t$) and to the base sample ($f_b$), and the tolerances $(\epsilon,\delta)$ that size the adaptive sample of each head from the base sample's variance estimate. In the released implementation the adaptive budget scales as $(\Phi^{-1}(1-\delta)/\epsilon)^2$, so only that ratio matters and we sweep $\epsilon{=}\delta$. The \vatt{} paper (App.~A.2, Table~3) chooses the parameters per target sparsity by a grid search on a few examples, keeping the configuration with the smallest local attention error among those that meet the target. We follow the same protocol under our accounting. For each target read we probe nominal budgets $\{0.4,0.6,0.8,1.0\}\times$target, sampling shares $\{0.25,0.5,0.75\}$ and $\epsilon{=}\delta\in\{0.025,0.05,0.1,0.2,0.3\}$ (plus, at 10\%, the points of the paper's own 10\% grid that fit a 10\% read) on four requests (two of qa\_1, two of niah\_multikey\_2), logging the realised read and the relative output error against dense attention at every layer call, and run the configuration with the lowest error among those whose realised read is within 5\% of the target on the full benchmark (Table~\ref{tab:vattprobe}). Tightening $\epsilon{=}\delta$ below 0.2 makes the adaptive budget overshoot the target at every nominal budget (at a 3\% target, $\epsilon{=}\delta{=}0.05$ realises 5.7--14\%), so at 0.5--5\% the protocol selects the loosest tolerance with the full nominal budget and a 50/50 split, which is the paper's default setting except for $\epsilon{=}\delta{=}0.3$ instead of $0.25$, and at 10\% it selects the paper's own grid point $f_t{=}7.5\%$, $f_b{=}2.5\%$. Table~\ref{tab:vattfixed} compares the protocol-chosen runs with the fixed $\epsilon{=}\delta{=}0.25$, 50/50 configuration on RULER-32K HARD; the \vatt{} rows of the other two models (Table~\ref{tab:multimodel}) use the fixed configuration. The \vatt{} paper reports 86.6 for \vatt{}(HashAttention) at ``10\% sparsity'' on the same tasks against 81.9 for HashAttention alone; under a strictly matched read the protocol recovers part of that gain (Table~\ref{tab:vattfixed}), and what remains is attributable to their search including tolerances down to $\epsilon{=}\delta{=}0.025$, whose adaptive budgets exceed a 10\% read under our accounting (Table~\ref{tab:vattprobe}), and to their fixed 128-token sink and local windows.
\begin{table}[h]
  \centering
  \caption{ \vatt{} parameter probe per target read (4 requests). For each $\epsilon{=}\delta$ the table lists the lowest-error configuration that meets the target, its realised read and its output error against dense attention; the chosen configuration is bold.}
  \label{tab:vattprobe}
  \setlength{\tabcolsep}{4pt}
  \begin{tabular}{lcc}
    \toprule
    configuration & realised read & output error \\
    \midrule
    \tabvattprobe
    \bottomrule
  \end{tabular}
\end{table}
\begin{table}[h]
  \centering
  \caption{ \vatt{} on RULER-32K HARD (mean accuracy), comparing the fixed $\epsilon{=}\delta{=}0.25$, 50/50 configuration with the protocol-chosen one of Table~\ref{tab:vattprobe}, with the measured read of the latter.}
  \label{tab:vattfixed}
  \begin{tabular}{lccc}
    \toprule
    target read & fixed ($\epsilon{=}\delta{=}0.25$, 50/50) & protocol-chosen & measured read \\
    \midrule
    \tabvattfixed
    \bottomrule
  \end{tabular}
\end{table}

\section{Question-free tasks}
\label{app:longbench}
\begin{table}[h]
  \centering
  \caption{ Beyond RULER at a 3\% read, on LongBench summarisation (no question chunk; ROUGE-L~\citep{lin2004rouge}, 30 requests/task) and LongBench-v2~\citep{bai2025longbenchv2} (0-shot accuracy, 60 requests). Absolute scores are low for dense attention too, a property of metric and benchmark, not of sparsity; \ours{} reads no question tokens. App.~\ref{app:longbench} gives details.}
  \label{tab:otherdata}
  \setlength{\tabcolsep}{3pt}
  \begin{tabular}{lccc}
    \toprule
    method (3\% read) & gov\_rep. & multi\_news & LB-v2 \\
    \midrule
    dense & 36.08 & 26.61 & 30.00 \\
    HashAttention & 35.34 & 22.09 & 23.30 \\
    \vatt{} & 31.05 & 16.33 & -- \\
    \textbf{\ours{} (static only)} & 35.53 & 26.80 & \textbf{33.30} \\
    \bottomrule
  \end{tabular}
\end{table}

This appendix expands Table~\ref{tab:otherdata} of the main text with all \ours{} variants on the question-free LongBench summarisation tasks (Table~\ref{tab:longbench}).
\begin{table}[h]
  \centering
  \caption{ LongBench summarisation tasks (no question chunk), 3\% read, 30 requests per task, ROUGE-L. ``static metric only'' uses no question at all (selector and partition from the calibrated $M$; the row of Table~\ref{tab:otherdata}); the last row treats the instruction tokens as the question.}
  \label{tab:longbench}
  
  \resizebox{\linewidth}{!}{%
    \begin{tabular}{lcc}
      \toprule
      method & gov\_report (ROUGE-L) & multi\_news (ROUGE-L) \\
      \midrule
      \tablongbenchrows
      \bottomrule
  \end{tabular}}
\end{table}

\paragraph{LongBench-v2.} As a non-synthetic check we also run LongBench-v2 (multiple choice, 0-shot prompt, accuracy) on its first 60 requests at 3\%, contexts truncated to 32000 tokens as everywhere else; the request count is small and the numbers are indicative only.
\begin{center}
  \begin{tabular}{lc}
    \toprule
    method (3\% total read) & LongBench-v2 0-shot accuracy \\
    \midrule
    \tablbvtwo
    \bottomrule
  \end{tabular}
\end{center}

\section{Composing with KV-cache compression}
\label{app:kvortho}
The three eviction compressors of Table~\ref{tab:kvcompress} follow the release of the PyramidKV paper (\texttt{KVCache-Factory}), which also ships the StreamingLLM and SnapKV baselines that paper compares against, and whose RULER driver fixes their hyper-parameters. The budget $b$ is an \emph{absolute} average cache size per layer rather than a fraction of the context, scores are max-pooled with kernel 7, and the retained window is the last $\alpha{=}8$ tokens for SnapKV and PyramidKV and $b-4$ for StreamingLLM, whose remaining four slots are the attention sinks of \citet{xiao2023streamingllm}. Every prefix key is scored by the attention that retained window pays to it, averaged over the query heads of its KV group, and the top of that score survives. StreamingLLM is the exception, since it consults no score and keeps tokens by position alone. PyramidKV's arithmetic allocation gives layer~0 a prefix budget of $2(b-\alpha)-(b-\alpha)/\beta$ and the last layer $(b-\alpha)/\beta$ with $\beta{=}20$, integer-divided steps in between, so the mean over layers is $b$; at $b{=}2048$ that is 3{,}986 entries in the first layer and 111 in the last. The prompt is encoded with \emph{full} attention and the compression applies to what is \emph{stored}, so only generation reads the compressed cache. That detail is not cosmetic, because biasing the selection call itself, which would encode the question chunk against the keys it has just chosen, costs 9.5 points on qa\_1 alone, even when the whole resident set is then read.

Both rows of each pair call the same selector, so a pair differs only in what the decode step reads. The difference from a baseline run is where the surviving set enters. Rather than adding the kept rows to the read mask, we hide the evicted rows from the attention itself, so they are invisible to the selector's scores, absent from the tail the partition clusters, and absent from the statistics of every cluster. Two consequences are worth stating. First, a cluster can lose most of its members to the eviction, and the accounting still charges its $1.5$ rows, so the read we charge is an upper bound on the read we take. Second, the partition is still built over all keys, not only the resident ones, so the $k$-means objective is slightly the wrong one, which costs accuracy rather than buying it.

\begin{table}[h]
  \centering
  \caption{ Table~\ref{tab:kvcompress} per task (RULER-32K HARD, Llama-3.1-8B-Instruct, 200 samples per task).}
  \label{tab:kvcompressfull}
  
  \setlength{\tabcolsep}{4pt}
  \begin{tabular}{lcccccccc}
    \toprule
    method & avg & qa\_1 & qa\_2 & vt & fwe & mk\_2 & mk\_3 & mv \\
    \midrule
    \tabkvcompress
    \bottomrule
  \end{tabular}
\end{table}

\section{Context-length sweep}
\label{app:length}
Table~\ref{tab:length} of \S\ref{sec:length} uses 100 samples per task at every length. The 16K and 32K sets are the public releases of the HashAttention and \vatt{} evaluation (\texttt{xAlg-AI/att-hub-ruler-16k} and \texttt{att-hub-ruler-32k}). The 32K column rescores the first 100 samples of existing runs of 200 samples with the same metric. For HashAttention and \vatt{} these are the runs of Table~\ref{tab:densityfull}, for \ours{} it is the run of Table~\ref{tab:multimodel}, and for \mpig{} it is the $K{=}10$ run described below. The 64K and 96K sets are generated with the RULER scripts~\citep{hsieh2024ruler} at maximum sequence lengths 65{,}536 and 98{,}304 tokens with the Llama-3.1 tokenizer, the base template and the same seven tasks (200 samples generated, the first 100 run), and the answer prefix is handed to the model exactly as in the public sets. Contexts are truncated to $L-768$ tokens at every length. All methods use the code paths of Table~\ref{tab:densityfull}, namely HashAttention with the released 64K signature weights, \vatt{} with the protocol configuration of App.~\ref{app:vattprotocol} for a 3\% read ($\epsilon{=}\delta{=}0.3$, 50/50 split, measured read reported), and \ours{} with the \qgs{}, the static$+$question partition and $C=64\cdot L/32768$ clusters with $E{=}0$, so that the proxy stream stays at 0.29\% of the read and a cluster holds about 500 tokens at every length. \mpig{} is recalibrated per length, because the realised density of a fixed $(K,L)$ shifts with the context (its $L{=}300$ configuration measures 3.20\% at 32K but 3.81\% at 16K and 3.41\% at 96K), so the rows use $K{=}10$ with $L{=}210$, $300$, $300$ and $250$ at 16K, 32K, 64K and 96K, whose measured reads are 2.99\%, 3.20\%, 3.19\% and 3.02\% (App.~\ref{app:magicpig}, whereas the density sweep of Table~\ref{tab:densityfull} uses $K{=}9$). Table~\ref{tab:lengthfull} lists the per-task scores and the reads.
\begin{table}[h]
  \centering
\caption{Table~\ref{tab:length} per task, with the total per-query read (measured for \vatt{}, \mpig{} and \ours{}).}
  \label{tab:lengthfull}
  \setlength{\tabcolsep}{3pt}
  \resizebox{\linewidth}{!}{%
    \begin{tabular}{llcccccccccc}
      \toprule
      context & method & read & avg & qa\_1 & qa\_2 & vt & fwe & mk\_2 & mk\_3 & mv \\
      \midrule
      \tablenrows
      \bottomrule
  \end{tabular}}
\end{table}

\end{document}